\documentclass{article}

\usepackage[main, final, nonatbib]{neurips_2026}

\usepackage[utf8]{inputenc} %
\usepackage[T1]{fontenc}    %
\usepackage{hyperref}       %
\usepackage{url}            %
\usepackage{booktabs}       %
\usepackage{amsfonts}       %
\usepackage{nicefrac}       %
\usepackage{microtype}      %
\usepackage{xcolor}         %
\usepackage{tikz}
\usepackage{tikz-cd}
\usepackage{multicol}
\usepackage{multirow}
\usepackage{amsmath}
\usepackage{algorithm}
\usepackage{algpseudocode}
\usepackage[table]{xcolor}
\usetikzlibrary{arrows.meta,calc,positioning,fit,decorations.pathreplacing}
\usepackage{amssymb}
\usepackage{pgfplots}
\usepackage{amsthm}
\newtheorem{proposition}{Proposition}
\pgfplotsset{compat=1.18}
\usetikzlibrary{calc}
\usepackage{xspace}
\usepackage{graphicx}
\algblockdefx[SUB]{Subroutine}{EndSubroutine}[1]{\textbf{subroutine} #1}{\textbf{end subroutine}}
\usepackage{placeins} %

\usepackage[normalem]{ulem}

\definecolor{darkgreen}{RGB}{0,100,0}

\definecolor{softblue}{RGB}{70,130,180}
\definecolor{softred}{RGB}{200,70,70}
\definecolor{softgreen}{RGB}{60,140,100}
\definecolor{softorange}{RGB}{230,140,60}
\definecolor{softpurple}{RGB}{140,110,180}
\definecolor{softteal}{RGB}{80,160,160}
\definecolor{softpink}{RGB}{220,120,150}
\definecolor{softgray}{RGB}{130,130,130}
\definecolor{softyellow}{RGB}{230,200,80}
\definecolor{softbrown}{RGB}{160,110,80}
\definecolor{softcyan}{RGB}{100,180,200}
\definecolor{softlime}{RGB}{140,180,70}
\definecolor{lightred}{RGB}{255,199,206}
\definecolor{darkred}{RGB}{220,80,80}

\newcommand{\ourmethod}{HeatKV\xspace}

\usepackage{subcaption}
\usepackage{enumitem}

\title{HeatKV: Head-tuned KV-cache Compression for Visual Autoregressive Modeling}

\author{
Jonathan Cederlund$^{1,2}$\thanks{Work completed during an internship at Arm.} \quad
Axel Berg$^{2}$ \quad
William Isaksson$^2$\thanks{Work completed while working at Arm.} \\
\textbf{Durmus Alp Emre Acar}$^{2}$ \quad
\textbf{Chuteng Zhou}$^{2}$ \quad
\textbf{Pontus Giselsson}$^{1}$ \\[1ex]
$^{1}$Dept.\ of Automatic Control, Lund University \quad
$^{2}$Arm \\
\texttt{jonathan.cederlund@gmail.com} \\
\texttt{\{axel.berg, william.isaksson, durmusalpemre.acar, chu.zhou\}@arm.com} \\
\texttt{pontus.giselsson@control.lth.se}
}

\begin{document}

\maketitle

\begin{abstract}
Visual Autoregressive (VAR) models have recently demonstrated impressive image generation quality while maintaining low latency. However, they suffer from severe KV-cache memory constraints, often requiring gigabytes of memory per generated image. We introduce HeatKV, a novel compression method that adapts cache allocation in each head based on its attention to previously generated scales. Using a small offline calibration set, the attention heads are ranked according to their attention scores over prior scales. Based on this ranking, we construct a static pruning schedule tailored to a given memory budget. Applied to Infinity-2B model, HeatKV achieves $2 \times$ higher compression ratio in memory allocation for KV cache compared to existing methods, while maintaining similar or better image fidelity, prompt alignment and human perception score. Our method achieves a new state-of-the-art (SOTA) for VAR model KV-cache compression, showcasing the effectiveness of fine-grained, head-specific cache allocation. Code and calibration script available at https://github.com/arm-research/heatkv.
\end{abstract}

\begin{figure}[h]
    \centering
    \setlength{\tabcolsep}{2pt}
    \renewcommand{\arraystretch}{0}

    \begin{tabular}{cccccc}

        \raisebox{1.68\height}{\rotatebox[origin=c]{90}{\textcolor{softred}{\textbf{Infinity 8B}}}} &
        \includegraphics[width=0.22\linewidth]{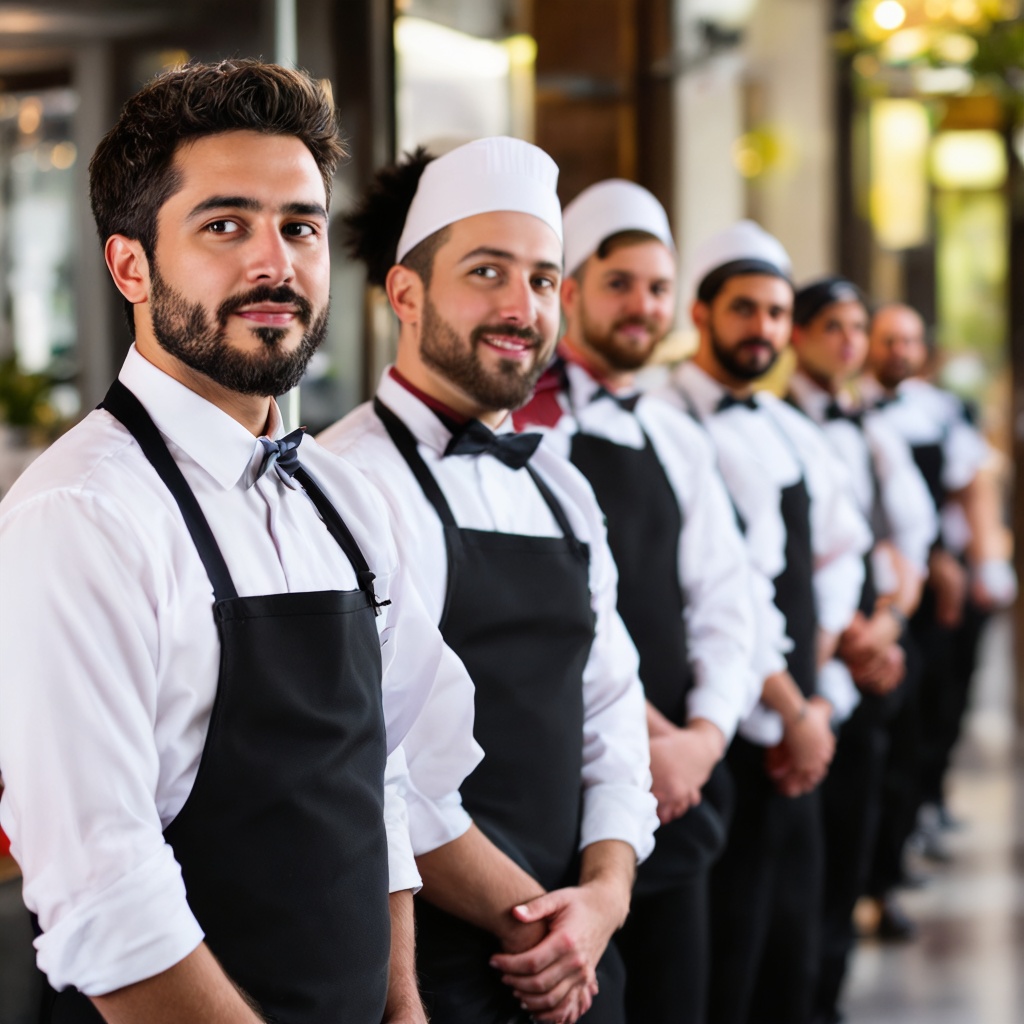} &
        \includegraphics[width=0.22\linewidth]{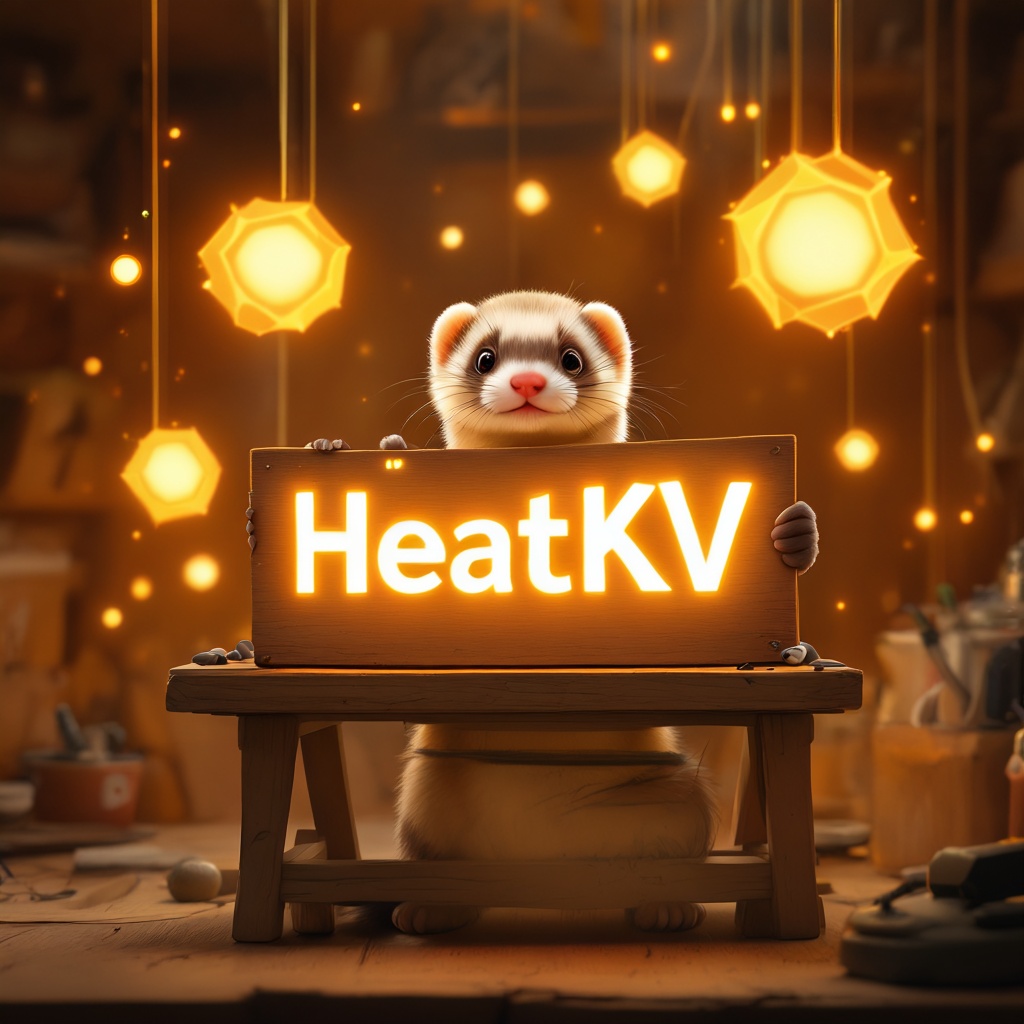} &
        \includegraphics[width=0.22\linewidth]{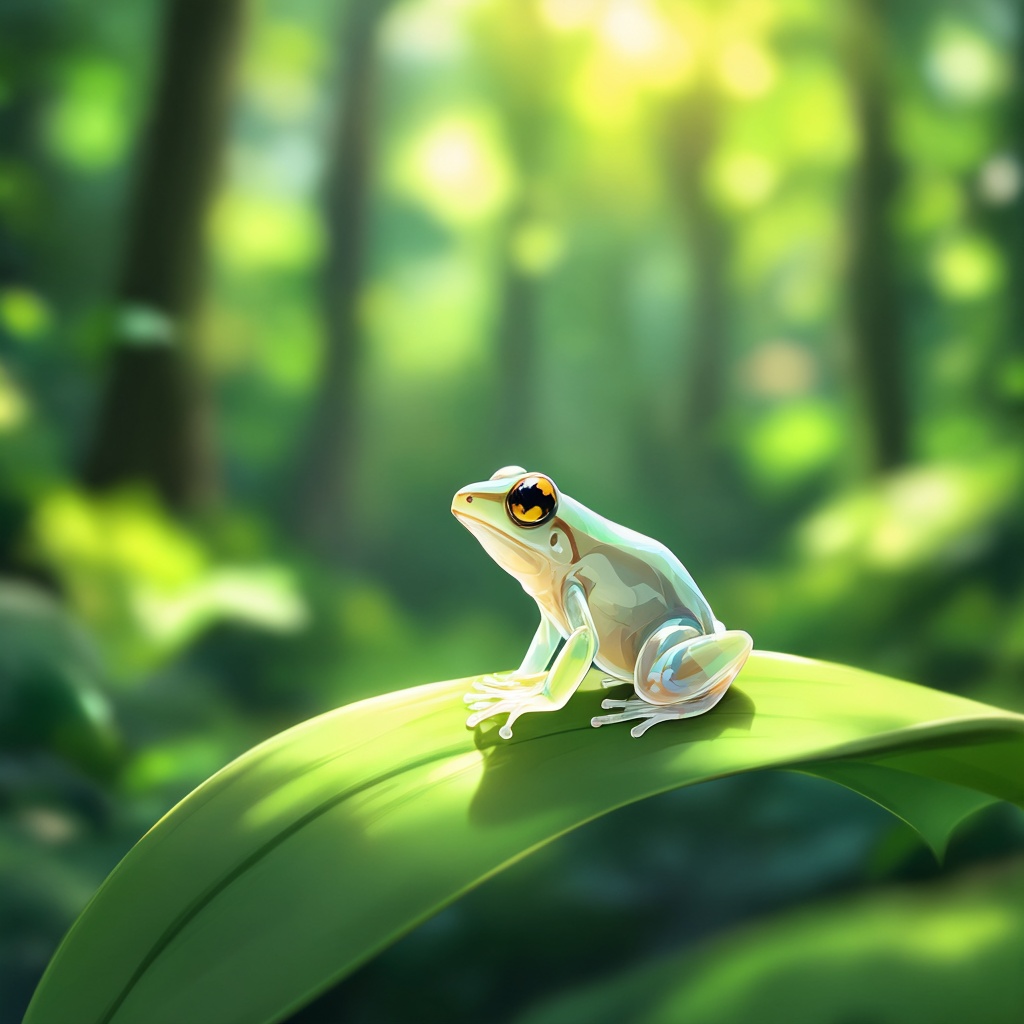} &
        \includegraphics[width=0.22\linewidth]{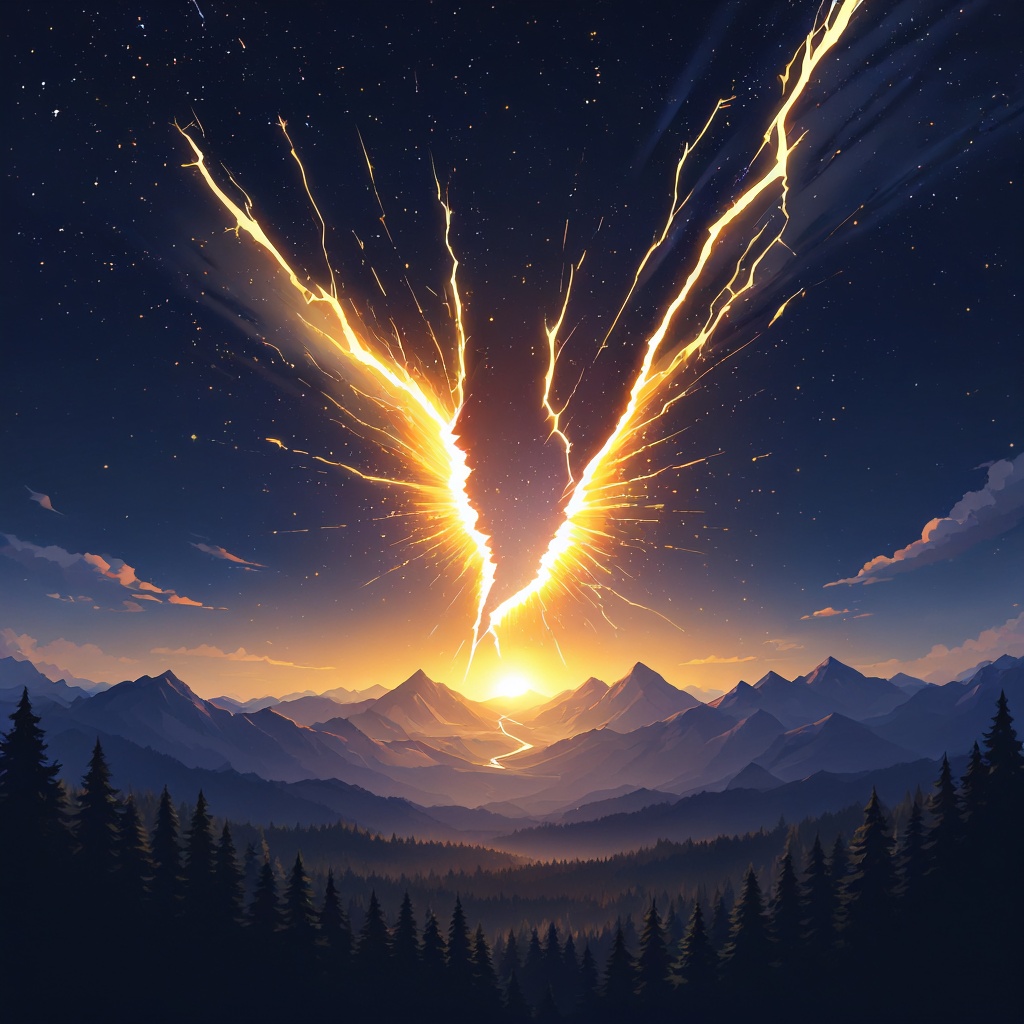} &
        \raisebox{1.08\height}{\rotatebox[origin=c]{90}{\textcolor{softred}{\textbf{42 GB KV cache}}}} \\[2pt]

        \raisebox{1.43\height}{\rotatebox[origin=c]{90}{\textcolor{softgreen}{\textbf{HeatKV}}}} &
        \includegraphics[width=0.22\linewidth]{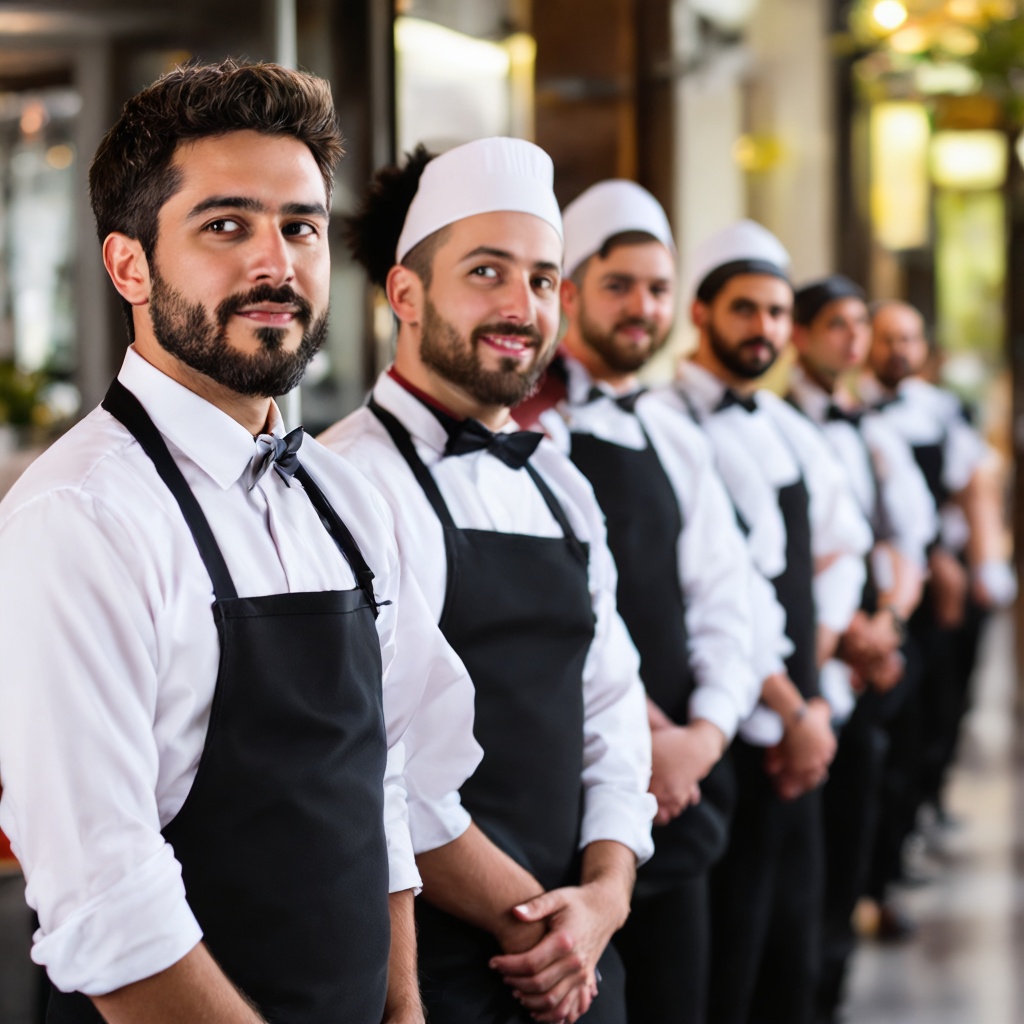} &
        \includegraphics[width=0.22\linewidth]{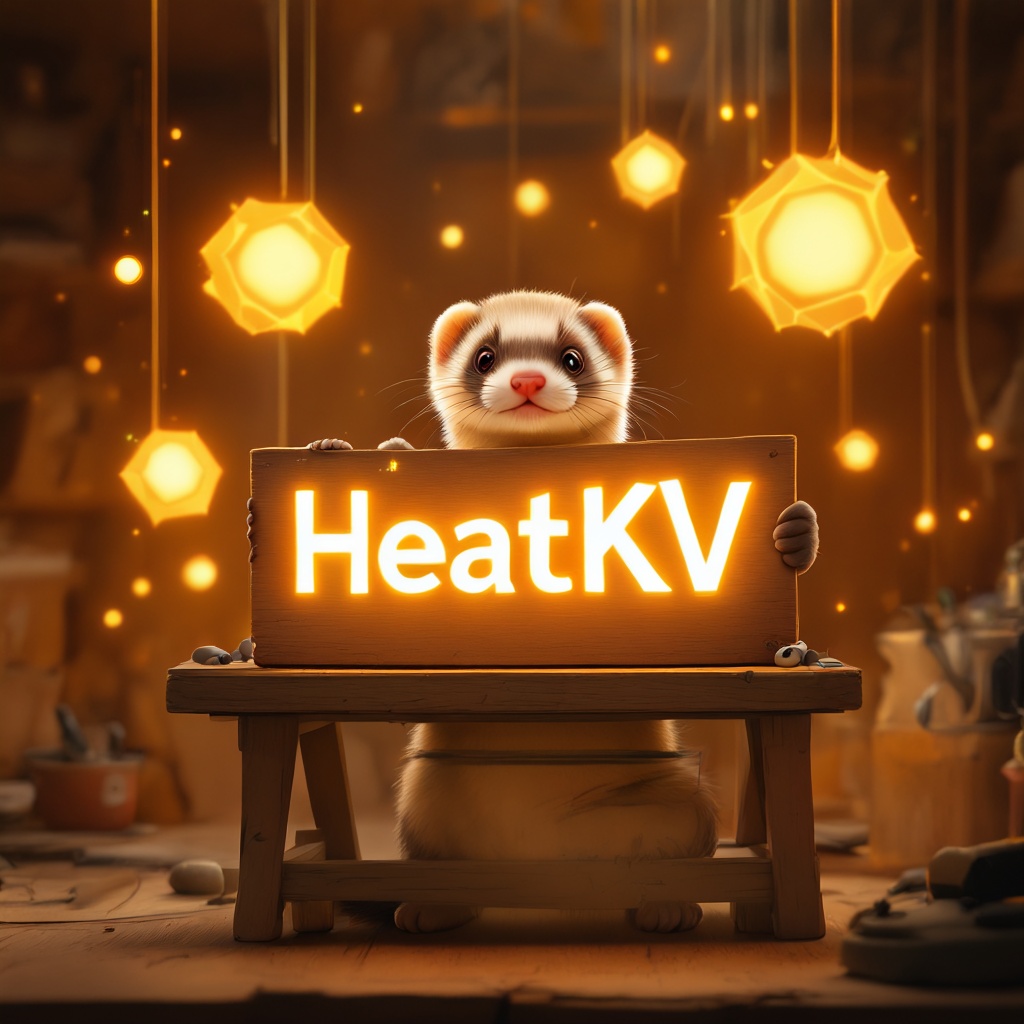} &
        \includegraphics[width=0.22\linewidth]{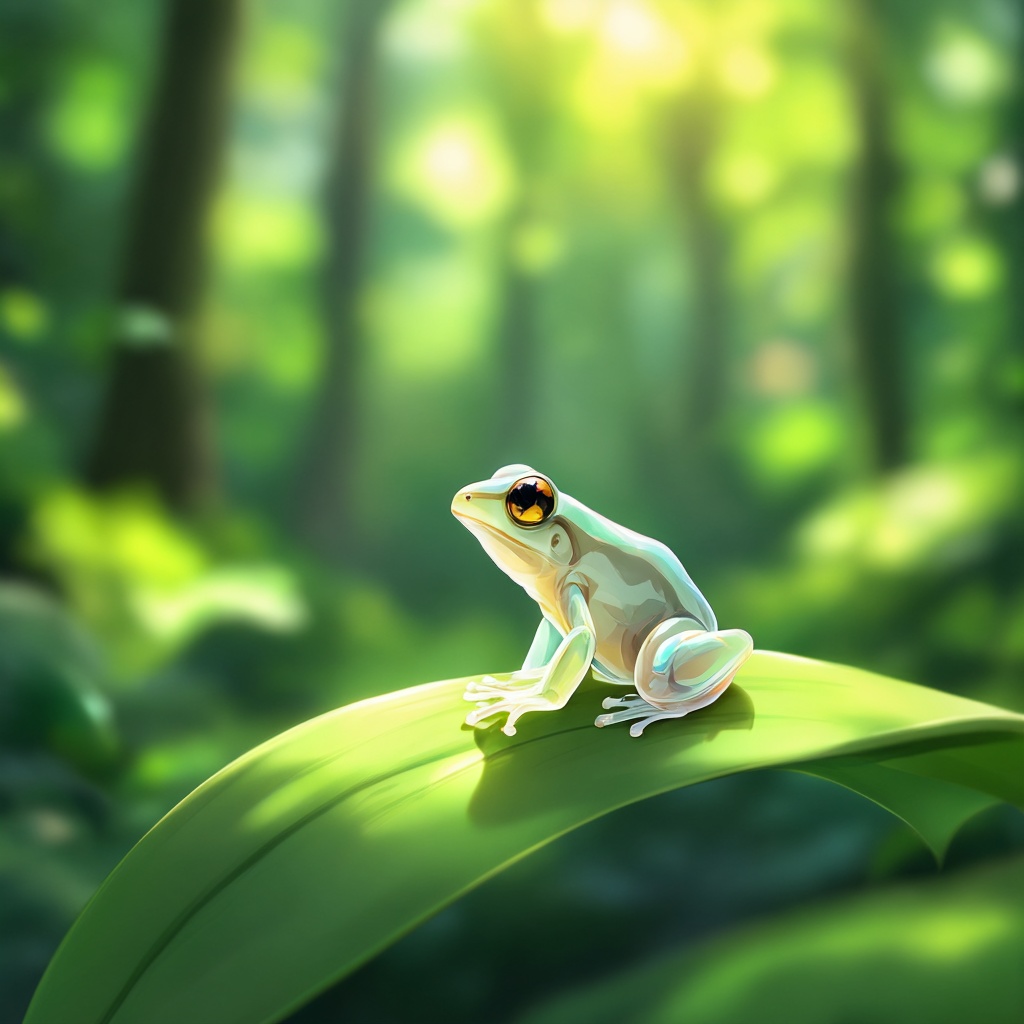} &
        \includegraphics[width=0.22\linewidth]{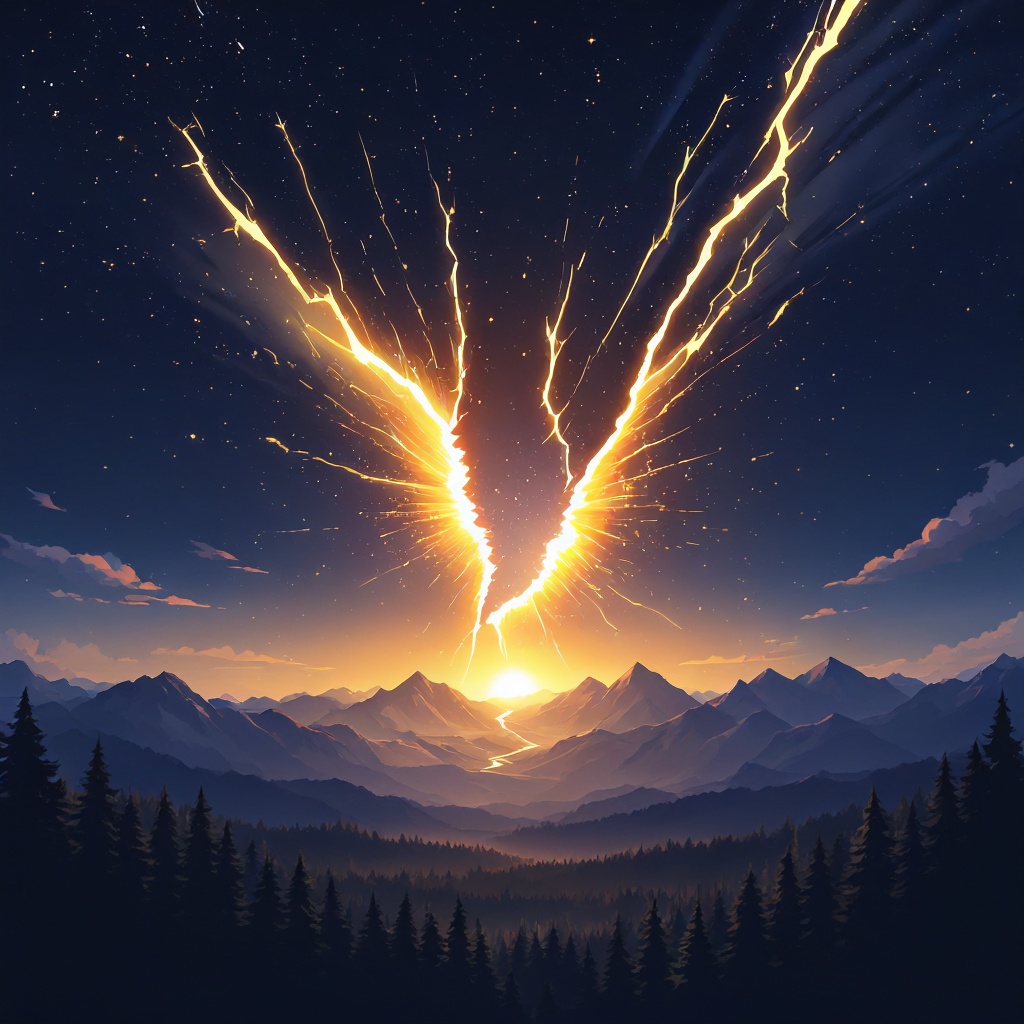} &
        \raisebox{1.04\height}{\rotatebox[origin=c]{90}{\textcolor{softgreen}{\textbf{2.1 GB KV cache}}}} 
    \end{tabular}

    \caption{Example images generated with Infinity-8B~\cite{han2025infinity} using \ourmethod at $20 \times $ compression ratio.}
    \label{fig:eight-image-grid}
\end{figure}

\section{Introduction}
\label{introduction}

As AI-assisted image generation has become more widespread in recent years, it poses new challenges for service providers. Since generative models are computationally expensive to run, specialized hardware accelerators need to be deployed at scale in order to serve multiple users at a low latency. For the service provider, there is an inherent trade-off to be made between latency, image quality and generation cost. As a consequence, there is a strong incentive for developing systems that can serve more users on the same infrastructure, while maintaining high image quality.

Among the popular model architectures for image generation are generative adversarial networks (GANs) \cite{goodfellow2014generative, karras2019style, sauer2022stylegan} and diffusion models \cite{dhariwal2021diffusion, rombach2022high, peebles2023scalable}. Recently, visual autoregressive (VAR) models have become an increasingly popular alternative due to their fast inference speed and scaling properties \cite{tian2024visual, han2025infinity}. By generating images autoregressively as a sequence of tokens at increasing resolution, it allows the models to exploit KV-caching to speed up the generation process. However, since the size of the KV-cache increases with the number of concurrent users, as well as with the image resolution, this increases the storage requirement for the accelerator's high bandwidth memory. State-of-the-art VAR models can require up to 10 GB of KV-cache for generating a single $1024 \times 1024$ image \cite{li2025scalekv}, which puts a hard constraint on the maximum batch size that can be used for generation. 

In order to decrease the memory requirement for VAR models, we propose \ourmethod (Head-tuned KV-cache Compression), a novel KV-cache compression strategy. \ourmethod leverages a small calibration dataset to extract attention patterns, enabling it to estimate the importance of each cache entry and selectively prune entries that receive little attention during generation. In contrast to previous methods, \ourmethod does not rely on classifying attention heads into distinct categories. Instead, we allow each head to assign a different budget to each image scale in the KV-cache, as shown in Figure \ref{fig:visual}, which provides finer-grained control over cache pruning, reducing wasted memory that would arise from coarse decisions. We summarize our main contributions as follows:

\begin{enumerate}[leftmargin=*]
    \item We observe that the attention patterns in VAR-models vary significantly between different scales within each attention head, but are relatively stable over input prompts. We therefore propose an offline budget allocation strategy that depends on how much the current feature map attends to prior scales within a given attention head. This allows each head to discard cache entries for non-important scales, while retaining the important ones.
    \item We develop a greedy KV-cache discarding algorithm that selects which cache entries to discard while making sure that the total cache memory stays below the assigned budget, by taking into account the increasing number of tokens generated in each scale. We show that this strategy is optimal under certain conditions.
    \item We provide thorough experimental validation of our method, showing state-of-the-art results across multiple models and benchmark metrics. Notably, our results show that we can achieve $2 \times$ higher compression ratio compared to previous methods using the Infinity-2B model \cite{han2025infinity}, while maintaining the same generated image quality on MS-COCO 2017.
\end{enumerate}

\begin{figure}[t]
    \centering
    \input{}
    \caption{Overview of \ourmethod. (1) VARs generate image in increasing scale size, causing growth to be superlinear; each head maintains a separate cache across scales. (2) Normalized attention scores over all calibration prompts are aggregated into a Scale-dependent Cumulative Attention Score (S-CAS)(Eq~\ref{eq:s-cas}), ranking scale importance per head (Alg~\ref{alg:strict-pruning}). (3) Given this ordering, low-importance head-scale pairs are pruned to meet a budget.}
    \label{fig:visual}
\end{figure}

\section{Related Work}
\label{relatedwork}
\paragraph{Autoregressive Image Generation} Multiple approaches to autoregressive image generation have been explored, including next pixel prediction \cite{chen2020generative} and next-token prediction \cite{esser2021taming, yu2022scaling, wang2025parallelized, sun2024autoregressive}. However, these approaches fail to adhere to the assumption of unidirectional dependency that underlies the autoregressive model, since nearby pixels (or patches of pixels) on a 2D grid have bidirectional correlation. Furthermore, the number of generation steps scales poorly with the image resolution. VAR models resolve these issues by instead forming a sequence of multi-scale token maps, where each token map corresponds to an image representation at a particular resolution \cite{tian2024visual}. Further developments of VAR models include hybrid diffusion approaches \cite{tang2025hart}, flow matching \cite{ren2025flowar} and multi-resolution generation \cite{jiao2026flexvar}. In addition, VAR models have been extended to other problem domains, such as monocular depth estimation \cite{el2025visual}, image super-resolution \cite{qu2025visual} and text-to-video generation \cite{liu2026infinitystar}.

\paragraph{KV-Cache Compression} Although many methods for KV-cache compression have been proposed in the context of large language models (LLMs) \cite{xiao2024efficient, zhang2023h2oheavyhitteroracleefficient, li2024snapkvllmknowslooking, feng2025adakv}, this topic is less mature in the context of VAR models, with a few notable exceptions \cite{li2025scalekv, qin2025headawarekvcachecompression, xu2026ams}. In contrast to LLMs, a VAR model generates multiple tokens in each autoregressive step, which requires a KV-cache eviction policy to discard up to half of its cache entries in each step in order to not exceed the memory budget.

ScaleKV \cite{li2025scalekv} classifies different layers in the transformer models into two distinct categories: drafters and refiners, which are assigned different memory budgets. The cache eviction policy is score-based, similar to \cite{li2024snapkvllmknowslooking}. HACK \cite{qin2025headawarekvcachecompression} takes a similar approach as ScaleKV, but instead of differentiating between transformer layers, it assigns cache budget at the attention head granularity. Each head is divided into one of two categories: structural or contextual, and depending on the category, the attention head is assigned a high or low memory budget. The two categories are also assigned different eviction policies. Finally, AMS-KV \cite{xu2026ams} divides layers into cache-demanding and cache-efficient layers. During generation, the budget for cache-demanding layers is dynamically expanded based on a similarity threshold. However, this makes the memory allocation input dependent and as a consequence the memory budget cannot be determined by the user before generation starts.

\section{Method}
\paragraph{Preliminaries}
\label{preliminaries}
VAR models generate images by next-scale prediction. At each generation step, an entire token map of increasing resolution is generated by conditioning on previous scales. Specifically, let $f \in \mathbb{R}^{h \times w \times C}$ be the image feature map at resolution $h \times w$ with $C$ channels, which is given by $f = \mathcal{E}(im)$, where $\mathcal{E}$ is a feature encoder and $im$ is the raw image. Discrete tokens $q \in [V]^{h \times w}$ are then obtained by quantizing the feature map as $q = \mathcal{Q}(f)$, where $V$ is the encoder codebook size and $\mathcal{Q}$ is a quantization function. The original VAR model used a VQVAE \cite{van2017neural} for tokenization. Recent work \cite{han2025infinity} has later shown that dimension-independent bitwise quantizers, such as lookup-free quantization \cite{yu2024language} or binary sphere quantization (BSQ) \cite{zhao2025image} allow for a more efficient scaling of the codebook size $V$.

In order to generate images autoregressively, the tokenization is performed at different resolutions (scales) $h_k \times w_k$ in order to obtain a sequence of token maps $(r_1, r_2, .., r_K)$, where $r_k \in [V]^{h_k \times w_k}$. The joint likelihood function at the final generation can then be decomposed as
\begin{equation}
    p(r_1, r_2, ..., r_K) = \prod_{k=1}^K p(r_k | r_1, ..., r_{k-1}, \Psi),
\end{equation}
where $K$ is the number of generation steps and $\Psi$ is the prompt embedding describing the final image.

\paragraph{Motivation}
\label{motivation}

\newlength{\imgw}
\setlength{\imgw}{0.32\linewidth}

\begin{figure}[t!]
  \centering
  \begin{minipage}[t]{\imgw}
    \centering
    \includegraphics[width=\linewidth]{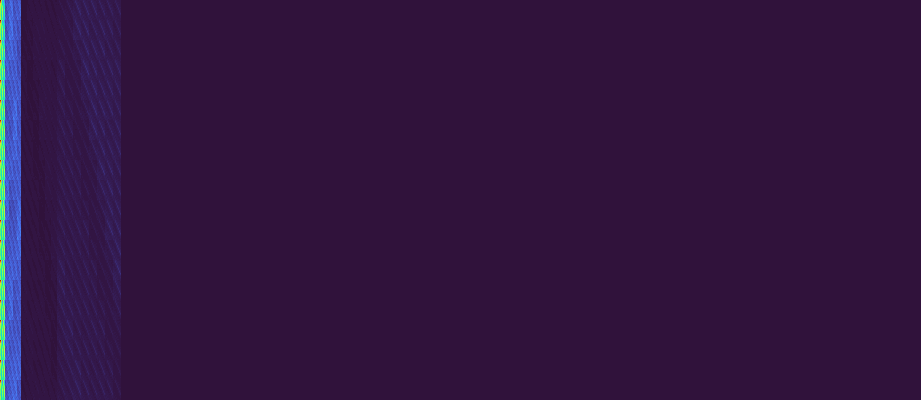}\\[-0.2em]
    {\tiny (a) Layer 1 Head 4}
  \end{minipage}\hfill
  \begin{minipage}[t]{\imgw}
    \centering
    \includegraphics[width=\linewidth]{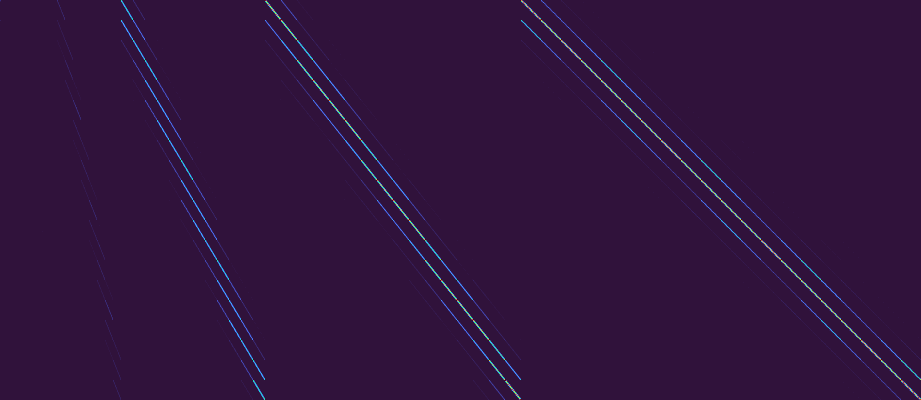}\\[-0.2em]
    {\tiny (b) Layer 22 Head 7}
  \end{minipage}\hfill
  \begin{minipage}[t]{\imgw}
    \centering
    \includegraphics[width=\linewidth]{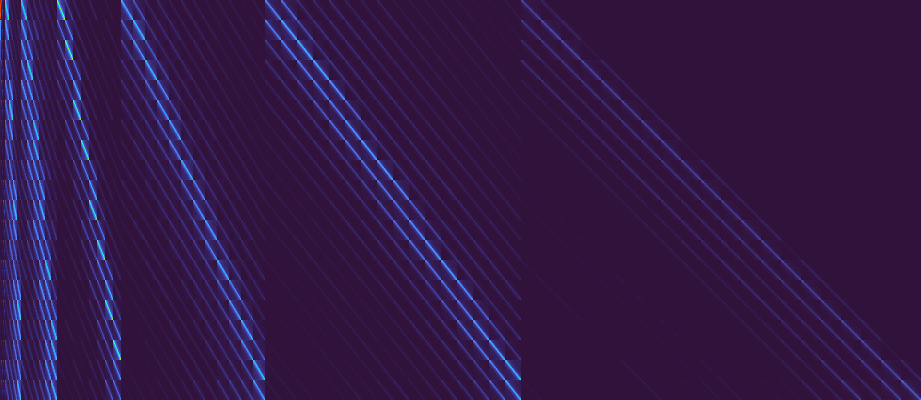}\\[-0.2em]
    {\tiny (c) Layer 1 Head 3}
  \end{minipage}\\[0.6em]
  \begin{minipage}[t]{\imgw}
    \centering
    \includegraphics[width=\linewidth]{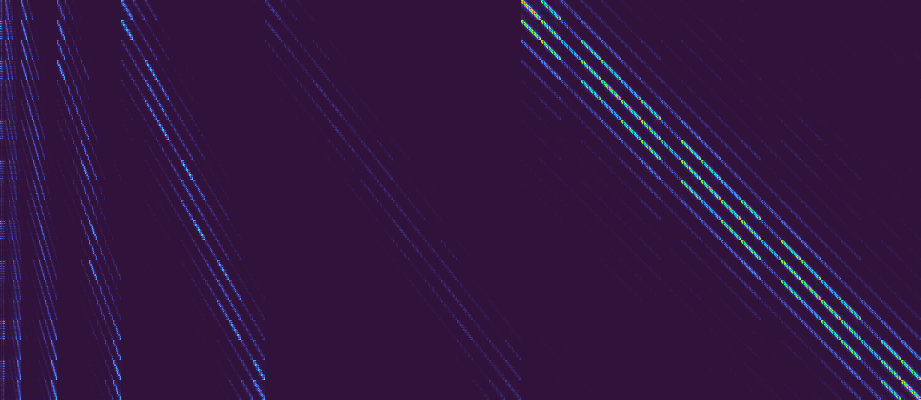}\\[-0.2em]
    {\tiny (d) Layer 3 Head 11}
  \end{minipage}\hfill
  \begin{minipage}[t]{\imgw}
    \centering
    \includegraphics[width=\linewidth]{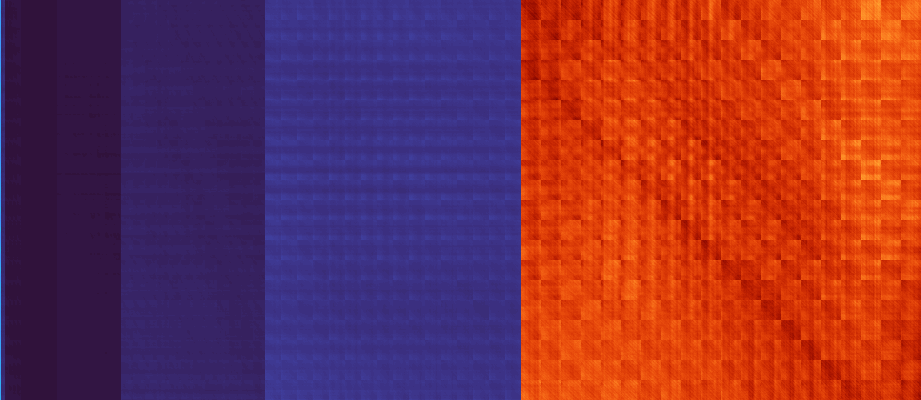}\\[-0.2em]
    {\tiny (e) Layer 1 Head 6}
  \end{minipage}\hfill
  \begin{minipage}[t]{\imgw}
    \centering
    \includegraphics[width=\linewidth]{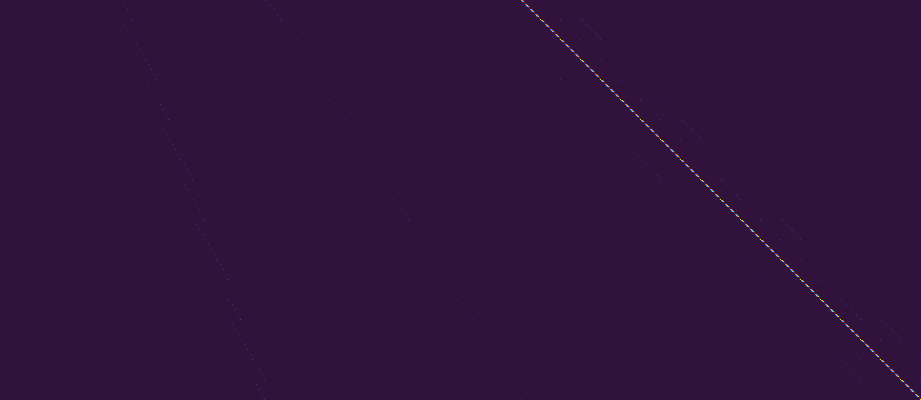}\\[-0.2em]
    {\tiny (f) Layer 2 Head 16}
  \end{minipage}

\caption{Different head-attention patterns. The y-axis shows tokens in the current scale (raster-ordered from the 2D image, with the top row as the first $w_k$ tokens). The x-axis shows cached tokens, from early scales (left) to the current scale (rightmost). (a) Early-scale attention, concentrating on scale 1. (b, c, d) Local cross-scale attention, (b) emphasizes recent scales, (c) focuses on early scales and (d) skips the previous scale to attend earlier ones. (e) Scale attention. (f) Strict self-attention.}%
\label{fig:attn_maps}
\end{figure}

Since each token map contains $h_k \times w_k$ tokens, the number of entries in the KV-cache grows at a super-linear rate with respect to the generation scale $k$. This makes generation at a high image resolution both memory consuming and computationally expensive, which motivates the use of KV-cache compression. The superlinear cache growth rate causes methods developed for LLMs to miss semantic structure and underperform on VARs. Previous methods developed for VARs, such as ScaleKV \cite{li2025scalekv} and HACK \cite{qin2025headawarekvcachecompression} show that attention patterns are stable within each head across prompts. Both methods leverage this stability to divide heads into two classes and apply a binary budget schedule (high-budget vs.\ low-budget heads). This strategy is effective, but as Figure \ref{fig:attn_maps} shows, it can be overly restrictive as heads in the same layer often require different cache budgets.

To address this, we propose \textbf{Head-tuned KV-cache compression (\ourmethod)} (Figure \ref{fig:visual}) with three components: head-scale importance, budget allocation and removal order. A head-scale denotes tokens from scale $k$ in head $h$. Unlike binary head classification, \ourmethod allows each head to retain an arbitrary subset of previous scales, controlling each head-scale pair independently. With $K$ the total number of scales to generate an image, each head can realize up to $2^K$ retention patterns. Following prior work \cite{xiao2024efficient, qin2025headawarekvcachecompression} we retain $s$ early \emph{sink scales} and never cache the last scale, resulting in $2^{(K - s - 1)}$ retention patterns. To allocate budget, we compute how many head-scales to prune after each generated scale, rank all head-scales by attention importance (their contribution to future predictions), and prune the lowest-ranked ones to meet the target $N_k$. Given the selected set, we greedily prune from later-layer heads to maximize short-term memory usage

\subsection{Head-Scale Importance}
For head ranking, previous methods \cite{li2025scalekv, qin2025headawarekvcachecompression} rely on variance-aware metrics. Instead we propose an attention probability based ranking over (layer, head, scale) tuples to determine budget allocation. Since each generation produces more than one token, we first define some scale-level quantities.

Let $t_k=w_k\times h_k$ be the number of tokens generated in scale $k$ and $c_k=\sum_{\tau=1}^{k}t_\tau$ be the total tokens generated including scale $k$. Given a calibration dataset $\mathcal{D}$ and a data point ${d\in \mathcal{D}}$, $\alpha^{d}_{\ell,h,k} \in {\mathbb{R}}^{t_k \times c_{k}}$ denotes attention probabilities from tokens at scale $k\in[1,K]$ to all generated tokens until the current scale $k$, for layer $\ell\in[1,L]$, head $h\in[1,H]$. We note that each row is softmax normalized, i.e.\, $\sum_{m=1}^{c_{k}}\alpha^d_{\ell,h,k}[n,m]=1$ for all $n\in[1,t_k]$ and $k\in[1,K]$.

To aggregate scale-level attention, we define scale based attention mass,  $\beta^{d}_{\ell,h}\in\mathbb{R}^{K\times K}$ corresponding to total attention of one scale to another scale. For each scale $k_1,k_2\in[1,K]$,
\begin{equation}
\beta^{d}_{\ell,h}[k_1,k_2] =
\begin{cases}
\frac{1}{t_{k_1}} \sum_{n=1}^{t_{k_1}} \sum_{m=c_{ k_2} - t_{k_2} +1}^{c_{ k_2}}
\alpha^{d}_{\ell,h,k_1}[n,m], & \text{if } k_2 \le k_1, \\
0, & \text{if } k_2 > k_1
\end{cases}
\end{equation}
where $m \in [c_{ k_2}-t_{ k_2}+1, c_{ k_2}]$ are the tokens at scale $k_2$. Lastly, we define $\beta_{\ell,h} = \frac{1}{|\mathcal{D}|}\sum_{d\in\mathcal{D}}\beta^{d}_{\ell,h}[k_1,k_2] \in \mathbb{R}^{K\times K}$ as the averaged scale-normalized attention mass over the calibration dataset. By definition, each $\beta[k_1]$ is a probability vector, i.e.\, $\sum_{k_2=1}^{K}\beta_{\ell,h}[k_1,k_2]=1$.

For per head ordering we propose \textbf{Cumulative Attention Score (CAS)} which offline creates a head-wise ordering of reliance on cached tokens. \textbf{CAS} measures the average attention mass by the tokens at the final scale $K$ to cached tokens (excluding the sink scales $s$) as
\begin{equation}
\mathrm{CAS}_{\ell,h} = \frac{1}{K-s}\sum_{\tau=s + 1}^{K-1} \beta_{\ell,h}[K,\tau],
\label{eq:cas}
\end{equation}
In order to determine the importance of each scale within a given head, we  introduce the \textbf{Scale-dependent Cumulative Attention Score (S-CAS)} (Eq.\ \ref{eq:s-cas}). S-CAS is constructed using the calibration dataset by aggregating attention mass from the future scales to a given (layer, head, scale) except the first $s$ sink scales. S-CAS represents how important (layer, head, scale) is for the future token generation. For a given layer $\ell\in[1,L]$, head $h\in[1,H]$ at scale $k\in[s+1,K-1]$,
\begin{equation}
\mathrm{S\text{-}CAS}_{\ell,h,k} = \frac{1}{K-k}\sum_{\tau = k + 1}^K  \beta_{\ell,h}[\tau,k],
\label{eq:s-cas}
\end{equation}
S-CAS defines a set of orders $\{O_i\ \colon s < i < K\}$ where each entry in the set orders all the heads based on their dependency on source-scale $i$ while generating future scales.

\subsection{Budget Allocation}
Given a KV-cache memory budget $b \in (0,1]$, we calculate how many heads have to be removed after each scale is generated. We follow Streaming-LLM \cite{xiao2024efficient} and always retain the first $s$ scales in the cache as an attention sink. Note that the input token when generating the first scale corresponds to the sentence embedding and it is therefore included in the early sink scales.

The model has $L$ self-attention layers with $H$ heads in each, resulting in $T = L \cdot H$ total heads. Let $N_k$ be the number of pruned heads after scale $k$. The total cache size after generating scale $k$ is then given by $ N_k \cdot c_s + (T - N_k) \cdot c_k $. Therefore, the minimal $N_k$ to remain under the budget constraint after generating scale $k$ is given by
\begin{equation} 
    N_k = \left\lceil T \cdot \dfrac{c_k - b \cdot c_{K-1}}{c_k - c_s} \right\rceil,
\end{equation}
for $k > s$ and 0 otherwise. As $N_k$ specifies the number of heads to prune, each source scale is removed in exactly $N_k$ heads, but not necessarily from the same heads. The exact set of head-scales to remove after scale $k$ is the prefix of length $N_k$ from the set of orderings defined by S-CAS. %

\subsection{Removal Order Optimization}

\begin{algorithm}[ht]
\caption{\colorbox{blue!30}{\textsc{Binary-\ourmethod}}, \colorbox{green!30}{\textsc{\ourmethod}}}
\label{alg:strict-pruning}
\begin{algorithmic}
\Require budget $b \in (0, 1]$, \colorbox{blue!30}{removal order $O$}, \colorbox{green!30}{removal orders $\{O_i \}_{i=s+1}^{K-1}$}, \# sink scales $s$, \\ \# heads in the model $T$, \# generation scales $K$, \# heads to remove after scale $k$ \{$N_k\}_{k=1}^{K-1}$ 
\vspace{0.5em}
\State $B \gets b \cdot T \cdot c_{K-1}$ \Comment{Maximum tokens in cache with budget b}
\State $A \gets \emptyset$ \Comment{Set of pruned heads}
\For{each scale index $k$}
    \State \colorbox{blue!30}{$G_k \gets \textsc{Prefix}(O, N_k)$} \Comment{Heads removed at scale $k$}
    \For{each prunable source scale $i$, $(s < i \le k)$}
        \State \colorbox{green!30}{$G_k \gets G_k \cup \textsc{Prefix}(O_i, N_{k})$} \Comment{Head-scales removed at scale $k$}
    \EndFor
    \State $A_k \gets A$ \Comment{Set of head\colorbox{green!30}{-scale}s removed before scale $k$}
    \Subroutine{\textsc{GreedyEarlyPruning}$(k, G_{k-1}, G_k, B)$}\label{line:earlyprune-start}
    \State $E_k \gets \emptyset$ \Comment{Set of early pruned heads}
    \State $R_k \gets \textsc{Sort}(G_k \setminus G_{k-1})$ by $(\mathrm{layer}\downarrow,$\colorbox{green!30}{$\mathrm{source\ scale}\downarrow,$}$ \mathrm{rank}\uparrow)$
    \For{layer $\ell = 1$ to $L$}
        \While{$\textsc{CacheSizeAfterLayer}(k, \ell, G_k, G_{k-1}, E_k) > B$}
            \State $E_k \gets E_k \cup \{\text{Pop}(R_k)\}$\Comment{Add next head\colorbox{green!30}{-scale}s to set of early removed}
        \EndWhile
    \EndFor
    \EndSubroutine\label{line:earlyprune-end}
    \State $A_k \gets A_k \cup E_k$, $A \gets G_k$
\EndFor
\State \Return $\{k \mapsto A_k\}$
\end{algorithmic}
\end{algorithm}

Previous methods \cite{li2025scalekv} enforce budgets at scale boundaries, (i.e., between forward passes), which can cause temporary KV-cache over-allocation when budgets are only adjusted between decoding steps. While a "soft" budget constraint only leads to negligible over allocation in LLMs (single-token generation), this becomes more significant in VAR models due to super-linear cache growth. We therefore introduce an optimized eviction scheme that enforces the budget constraint every time tokens are added to the cache, avoiding transient budget violations. This may require pruning some heads one scale earlier (“early pruning”). One simple policy to remain under the budget at all times is to prune before a scale and not after. While this naive method remains under budget, it does not maximize the budget utilization. To improve budget utilization we propose Binary-\ourmethod for the binary case and \ourmethod for the scale-dependent case in Algorithm \ref{alg:strict-pruning}.

\paragraph{Binary-\ourmethod}
In the binary setting, a head either retains all non-sink cached scales or only the sink scales, so the prunable object is a head $(\ell,h)$. Let
\[
G_k = \operatorname{Prefix}(O, N_k)
\]
denote the heads that must be in sink-only mode by the end of scale $k$, where $O$ ranks heads from least to most dependent on cached tokens using CAS. Since $N_k$ is non-decreasing, we have $G_{k-1} \subseteq G_k$, and the newly pruned heads at scale $k$ are $G_k \setminus G_{k-1}$. Under a naive boundary-only policy, these heads could be evicted after scale $k$. In our strict policy, however, the cache budget $B$ must hold after every layer. $E_k$ is defined by \textsc{GreedyEarlyPruning} in Algorithm~\ref{alg:strict-pruning}, the greedy subset of early pruned heads that must be evicted before scale $k$ starts to remain under budget. Define
\[
A_k = G_{k-1} \cup E_k
\]
as the set of heads absent from the cache at the beginning of scale $k$. The remaining heads in $G_k \setminus A_k$ are evicted immediately after their layer has executed. Algorithm~\ref{alg:strict-pruning} computes $E_k$ by simulating the cache size layer by layer and adding heads from $G_k \setminus G_{k-1}$ whenever \textsc{CacheSizeAfterLayer} (Appendix~\ref{appendix:algorithms}) exceeds $B$. Candidate heads are selected in decreasing layer index and, within the same layer, increasing rank in $O$. By Proposition~\ref{prop:greedy-optimality}, \textsc{GreedyEarlyPruning} in Algorithm~\ref{alg:strict-pruning} yields a minimum-cardinality feasible early-pruning set. By minimizing the cardinality, it minimizes the number of tokens we prune early thereby maximizing the utilization of the budget. Figure~\ref{fig:removal_order_binary} visualizes the roles of $G_k$, $E_k$, and $A_k$.

\begin{proposition}[Optimality of GreedyEarlyPruning in Binary-HeatKV]
Let \(\mathcal{E}_k\) be any subset of \(G_k \setminus G_{k-1}\), an early-pruning set such that the cache usage at scale \(k\) and layer \(\ell\) does not exceed a budget \(B\) for all \(\ell\).
\[
\mathcal{E}_k
\in \left\{
E \subseteq G_k \setminus G_{k-1}
\;\middle|\;
f_{\mathrm{past}}^{(k)}(\ell)
+ f_{\mathrm{future}}^{(k)}(\ell)
- f_{\mathrm{early}}^{(k)}(\ell;E)
\le B
\;\;\forall \ell
\right\}.
\]
 Here, \(f_{\mathrm{past}}^{(k)}\) and \(f_{\mathrm{future}}^{(k)}\) denote the cache from past and future layers at scale \(k\), respectively, and \(f_{\mathrm{early}}^{(k)}\) is the reduction induced by \(\mathcal{E}_k\); all terms are non-negative. Each early pruned head \(h\) reduces the cache size for all layers until its layer. Thereby, pruning from a layer will reduce the cache by at least as much as an earlier head would for all layers $\ell$. Then selecting heads greedily, as in the binary case for \textsc{GreedyEarlyPruning} in Algorithm~\ref{alg:strict-pruning}, decreasing layer order until feasibility is reached yields the minimum-cardinality set \(E_k\). The proof is given in Appendix \ref{appendix:proof}. 
\label{prop:greedy-optimality}
\end{proposition}

\paragraph{\ourmethod}
In the scale-dependent setting, the prunable unit is a head-scale $(i,\ell,h)$, where $i$ denotes the source scale and $(\ell,h)$ the head. For each non-sink source scale $i$, S-CAS defines an ordering $O_i$ ranking head from least to most dependent on tokens from $i$. Following Algorithm~\ref{alg:strict-pruning}, after scale $k$ we form $G_k$ by taking the prefix of length $N_k$ from each $O_i$, $(i \leq k)$ and uniting them. Thus, $G_k$ is the set that must be absent by the end of scale $k$, and $G_k \setminus G_{k-1}$ are the newly pruned elements. To enforce the budget, some head-scales may need to be evicted early: $E_k \subseteq G_k \setminus G_{k-1}$, with $A_k = G_{k-1} \cup E_k$ as the set absent at the beginning of scale $k$. The remaining elements in $G_k \setminus A_k$ are evicted immediately after their layer executes. The greedy order now sorts candidates by decreasing layer index, then decreasing source-scale index, and finally increasing rank within $O_i$. The additional source-scale tie-break favors later scales as they free more memory. See \ourmethod in Algorithm~\ref{alg:strict-pruning}.

\providecommand{\gridcellsize}{0.12cm}

\providecommand{\DrawHeadLayerGrid}[6]{%
    \begin{tikzpicture}[x=\gridcellsize,y=\gridcellsize,baseline=(current bounding box.north)]
        \path[use as bounding box] (0,-1) rectangle (16,32);
        
        \if\relax\detokenize{#4}\relax
        \else
            \edef\FilledCoordinates{#4}
            \foreach \head/\layer in \FilledCoordinates {
                \pgfmathtruncatemacro{\yinvert}{31-\layer}
                \fill[#3] (\head,\yinvert) rectangle ++(1,1);
            }
        \fi

        \draw[step=1, very thin, draw=gray!45] (0,0) grid (16,32);
        \draw[thick] (0,0) rectangle (16,32);

        \ifnum#5=1
            \foreach \x in {1,6,...,16} {
                \node[font=\tiny, anchor=north] at (\x-0.5,-0.5) {\x};
            }
        \fi
        
        \ifnum#6=1
            \foreach \y in {1,8,16,24,32} {
                \pgfmathtruncatemacro{\ypos}{32-\y}
                \node[font=\tiny, anchor=east] at (-0.8,\ypos+0.5) {\y};
            }
        \fi

        \node[font=\scriptsize, align=center, anchor=center] at (8,36.2) {#1};
        \node[font=\scriptsize, align=center, anchor=center] at (8,33.6) {#2};
    \end{tikzpicture}%
}

\def\ScaleSevenLowBudget{%
11/0, 12/1, 15/1, 9/1, 14/1, 8/1, 0/1, 4/1, 11/1, 2/1, 7/1, 12/2, 8/2, 3/2, 14/3, 2/3, 7/3, 1/3, 10/3, 14/4, 0/5, 2/5, 14/5, 11/5, 13/5, 6/5, 8/6, 10/6, 2/6, 15/6, 0/6, 14/7, 10/7, 2/7, 7/7, 0/8, 12/8, 6/8, 13/8, 8/8, 9/9, 6/9, 4/9, 14/9, 0/9, 8/9, 14/10, 6/10, 15/10, 1/10, 12/10, 3/10, 2/10, 13/10, 4/10, 8/10, 9/10, 7/10, 10/10, 11/11, 5/11, 6/11, 12/11, 8/11, 14/11, 4/11, 9/11, 3/11, 7/11, 13/11, 9/12, 6/12, 7/12, 11/12, 4/12, 5/12, 8/12, 15/12, 2/12, 0/12, 13/12, 1/12, 7/13, 8/13, 1/13, 0/13, 11/13, 13/13, 12/14, 13/14, 0/14, 2/14, 14/14, 1/14, 8/15, 14/15, 13/15, 3/15, 15/15, 15/16, 2/16, 11/16, 8/16, 6/17, 15/17, 12/17, 9/18, 11/18, 3/18, 0/18, 2/18, 13/19, 8/19, 10/20, 2/20, 8/20, 11/20, 13/20, 12/20, 9/20, 0/20, 14/20, 8/21, 1/21, 4/21, 15/21, 5/21, 9/21, 2/21, 3/21, 12/21, 10/22, 6/24, 11/25, 12/26, 5/26, 9/26, 7/26, 8/27, 9/27, 6/27, 15/28, 7/28, 14/28, 2/28, 9/28, 1/28, 0/28, 6/28, 13/28, 2/29, 8/29, 14/29, 9/29, 3/29, 5/30, 12/30, 0/30, 10/30
}

\def\ScaleEightLowBudget{%
11/0, 12/1, 15/1, 9/1, 14/1, 8/1, 0/1, 4/1, 11/1, 2/1, 7/1, 10/1, 12/2, 8/2, 3/2, 7/2, 13/2, 14/3, 2/3, 7/3, 1/3, 10/3, 13/3, 15/3, 6/3, 12/3, 14/4, 2/4, 0/4, 10/4, 11/4, 0/5, 2/5, 14/5, 11/5, 13/5, 6/5, 12/5, 5/5, 8/5, 7/5, 8/6, 10/6, 2/6, 15/6, 0/6, 6/6, 4/6, 7/6, 3/6, 14/7, 10/7, 2/7, 7/7, 5/7, 4/7, 0/7, 1/7, 3/7, 9/7, 15/7, 0/8, 12/8, 6/8, 13/8, 8/8, 3/8, 10/8, 1/8, 7/8, 15/8, 5/8, 14/8, 4/8, 9/9, 6/9, 4/9, 14/9, 0/9, 8/9, 3/9, 15/9, 13/9, 7/9, 14/10, 6/10, 15/10, 1/10, 12/10, 3/10, 2/10, 13/10, 4/10, 8/10, 9/10, 7/10, 10/10, 0/10, 11/11, 5/11, 6/11, 12/11, 8/11, 14/11, 4/11, 9/11, 3/11, 7/11, 13/11, 1/11, 15/11, 10/11, 2/11, 9/12, 6/12, 7/12, 11/12, 4/12, 5/12, 8/12, 15/12, 2/12, 0/12, 13/12, 1/12, 14/12, 12/12, 7/13, 8/13, 1/13, 0/13, 11/13, 13/13, 3/13, 10/13, 2/13, 12/13, 15/13, 12/14, 13/14, 0/14, 2/14, 14/14, 1/14, 10/14, 4/14, 3/14, 11/14, 9/14, 15/14, 8/15, 14/15, 13/15, 3/15, 15/15, 12/15, 2/15, 1/15, 11/15, 5/15, 10/15, 9/15, 15/16, 2/16, 11/16, 8/16, 10/16, 9/16, 12/16, 14/16, 3/16, 7/16, 6/16, 6/17, 15/17, 12/17, 4/17, 7/17, 1/17, 5/17, 14/17, 13/17, 9/17, 9/18, 11/18, 3/18, 0/18, 2/18, 15/18, 1/18, 14/18, 4/18, 13/18, 7/18, 13/19, 8/19, 6/19, 2/19, 0/19, 3/19, 15/19, 12/19, 9/19, 10/19, 5/19, 11/19, 4/19, 10/20, 2/20, 8/20, 11/20, 13/20, 12/20, 9/20, 0/20, 14/20, 15/20, 4/20, 3/20, 7/20, 6/20, 5/20, 1/20, 8/21, 1/21, 4/21, 15/21, 5/21, 9/21, 2/21, 3/21, 12/21, 13/21, 11/21, 0/21, 14/21, 10/22, 0/22, 2/22, 4/22, 13/22, 1/22, 9/22, 6/22, 5/22, 8/23, 6/24, 5/24, 11/25, 0/25, 12/26, 5/26, 9/26, 7/26, 15/26, 4/26, 8/27, 9/27, 6/27, 4/27, 11/27, 14/27, 12/27, 0/27, 15/28, 7/28, 14/28, 2/28, 9/28, 1/28, 0/28, 6/28, 13/28, 3/28, 8/28, 10/28, 2/29, 8/29, 14/29, 9/29, 3/29, 10/29, 7/29, 5/29, 5/30, 12/30, 0/30, 10/30, 15/30, 9/30, 0/31, 15/31, 11/31, 5/31, 14/31, 9/31, 13/31
}

\def\ScaleSevenFinal{}
\def\ScaleEightFinal{
11/0, 12/1, 15/1, 9/1, 14/1, 8/1, 0/1, 4/1, 11/1, 2/1, 7/1, 12/2, 8/2, 3/2, 14/3, 2/3, 7/3, 1/3, 10/3, 14/4, 0/5, 2/5, 14/5, 11/5, 13/5, 6/5, 8/6, 10/6, 2/6, 15/6, 0/6, 14/7, 10/7, 2/7, 7/7, 0/8, 12/8, 6/8, 13/8, 8/8, 9/9, 6/9, 4/9, 14/9, 0/9, 8/9, 14/10, 6/10, 15/10, 1/10, 12/10, 3/10, 2/10, 13/10, 4/10, 8/10, 9/10, 7/10, 10/10, 11/11, 5/11, 6/11, 12/11, 8/11, 14/11, 4/11, 9/11, 3/11, 7/11, 13/11, 9/12, 6/12, 7/12, 11/12, 4/12, 5/12, 8/12, 15/12, 2/12, 0/12, 13/12, 1/12, 7/13, 8/13, 1/13, 0/13, 11/13, 13/13, 12/14, 13/14, 0/14, 2/14, 14/14, 1/14, 8/15, 14/15, 13/15, 3/15, 15/15, 15/16, 2/16, 11/16, 8/16, 6/17, 15/17, 12/17, 9/18, 11/18, 3/18, 0/18, 2/18, 13/19, 8/19, 10/20, 2/20, 8/20, 11/20, 13/20, 12/20, 9/20, 0/20, 14/20, 8/21, 1/21, 4/21, 15/21, 5/21, 9/21, 2/21, 3/21, 12/21, 10/22, 6/24, 11/25, 12/26, 5/26, 9/26, 7/26, 8/27, 9/27, 6/27, 15/28, 7/28, 14/28, 2/28, 9/28, 1/28, 0/28, 6/28, 13/28, 2/29, 8/29, 14/29, 9/29, 3/29, 5/30, 12/30, 0/30, 10/30, 0/31, 15/31, 11/31, 5/31, 14/31, 9/31, 13/31, 15/30, 9/30, 10/29, 7/29, 5/29, 3/28, 8/28, 10/28, 4/27, 11/27, 14/27, 12/27, 0/27, 15/26
}
\def\ScaleSevenActivation{}
\def\ScaleEightActivation{0/31, 15/31, 11/31, 5/31, 14/31, 9/31, 13/31, 15/30, 9/30, 10/29, 7/29, 5/29, 3/28, 8/28, 10/28, 4/27, 11/27, 14/27, 12/27, 0/27, 15/26}

\begin{figure}[t]
    \centering
    \begin{tikzpicture}
    
        \node[inner sep=0pt] (grid1) at (0,0) {\DrawHeadLayerGrid{Scale 7}{$G_7$}{red!80}{\ScaleSevenLowBudget}{1}{1}};
        \node[inner sep=0pt, right=0.2cm of grid1] (grid2) {\DrawHeadLayerGrid{Scale 7}{$A_7$}{red!80}{\ScaleSevenFinal}{0}{0}};
        \node[inner sep=0pt, right=0.2cm of grid2] (grid3) {\DrawHeadLayerGrid{Scale 7}{$E_7$}{orange!75}{\ScaleSevenActivation}{0}{0}};
        \node[inner sep=0pt, right=0.4cm of grid3] (grid4) {\DrawHeadLayerGrid{Scale 8}{$G_8$}{red!80}{\ScaleEightLowBudget}{0}{0}};
        \node[inner sep=0pt, right=0.2cm of grid4] (grid5) {\DrawHeadLayerGrid{Scale 8}{$A_8$}{red!80}{\ScaleEightFinal}{0}{0}};
        \node[inner sep=0pt, right=0.2cm of grid5] (grid6) {\DrawHeadLayerGrid{Scale 8}{$E_8$}{orange!75}{\ScaleEightActivation}{0}{0}};

        \node[font=\normalsize, rotate=90, text=gray!80] (seperator) at ($ (grid3.east)!0.5!(grid4.west) + (0,-0.1) $) {- - - - - - - - -};

        \node[font=\scriptsize] at ($ (grid1.south) + (0,-0.2) $) {head index};
        \node[font=\scriptsize, rotate=90] at ($ (grid1.west) + (-0.5,0) $) {layer index};
        \node[
            anchor=north,
        ] at ($ (seperator.south) + (0,-1.67) $)
        {
        \begin{tikzpicture}[x=0.25cm,y=0.25cm]
            \node[
                draw,
                fill=red!80,
                minimum size=0.3cm,
                inner sep=0pt
            ] (legend_pruned) at (0.5,0.5) {};
            \node[right=0.2cm,
    text height=1.5ex,
    text depth=0.25ex] at (legend_pruned.east) {\scriptsize Pruned head};
            
            \node[
                draw,
                fill=white,
                minimum size=0.3cm,
                inner sep=0pt,
                right=1.8cm
            ] (legend_full) at (legend_pruned.east) {};
            \node[right=0.2cm,
    text height=1.5ex,
    text depth=0.25ex] at (legend_full.east) {\scriptsize Full cache};
            
            \node[
                draw,
                fill=orange!75,
                minimum size=0.3cm,
                inner sep=0pt,
                right=1.8cm
            ] (legend_early) at (legend_full.east) {};
            \node[right=0.2cm,
    text height=1.5ex,
    text depth=0.25ex] at (legend_early.east) {\scriptsize Pruned early};
        \end{tikzpicture}
        };
    \path[use as bounding box] (-1,1) rectangle (20,2.7);
    \end{tikzpicture}
    \caption{Removal order using \textsc{Binary-\ourmethod} with \textsc{GreedyEarlyPruning} in Algorithm \ref{alg:strict-pruning} on Infinity-2B, 10\% budget. Each small square is an attention head, specified by layer and head index.}
    \label{fig:removal_order_binary}
\end{figure}


\label{method}

\section{Experimental Results}

\paragraph{Base Models} For experiments, we use the Infinity-2B and 8B models \cite{han2025infinity} as our baselines. To study robustness under different compression levels, we retain 4\%, 10\%, and 20\% of the original KV-cache budget. We compare our method to four other KV-cache compression baselines. StreamingLLM \cite{xiao2024efficient} retains a set of early tokens (sink) and the most recent tokens. SnapKV \cite{li2024snapkvllmknowslooking} retains tokens based on attention scores. HACK \cite{qin2025headawarekvcachecompression} and ScaleKV \cite{li2025scalekv} use offline binary classification to assign budgets per head and per layer. For \ourmethod and HACK we use $N = 10$ calibration prompts generated by Qwen3-VL-2B-Instruct \cite{qwen3technicalreport}. These 10 prompts are used for the binary head classification in HACK and to calculate S-CAS for \ourmethod. Furthermore, we use $s = 3$ sink scales for StreamingLLM, HACK and HeatKV. Experiments for the Infinity-2B model as well as the latency measurements are performed on an NVIDIA H100 GPU, while Infinity-8B results are generated on an NVIDIA B200 GPU. For all experiments we use classifier-free guidance $cfg = 3$ and a default batch size of 8. Both the default resolution of 1024 by 1024 and the default scale schedule (1,2,4,6,8,12,16,20,24,32,40,48,64) are used on both models.

\paragraph{Datasets and Metrics}  We evaluate cache compression from three complementary perspectives: fidelity, prompt alignment, and perceptual quality. For fidelity, we use the MS-COCO 2017 validation set \cite{lin2015microsoftcococommonobjects}, (5k images) and compare compressed outputs to the base model using Peak Signal-to-Noise Ratio (PSNR), Learned Perceptual Image Patch Similarity (LPIPS) \cite{lpips}, and Fr\'echet Inception Distance (FID) \cite{heusel2018ganstrainedtimescaleupdateFID}. For compositional alignment, we use GenEval \cite{geneval}, which measures properties such as object count, spatial position, color attributes, and object co-occurrence. This benchmark provides a direct measure of how well a compressed model preserves prompt-conditioned reasoning. For perceptual quality, we report Human Preference Score v2.1 (HPSv2.1) \cite{wu2023human}, a CLIP-based predictor of human preference. Together, these benchmarks measure whether KV-cache compression preserves the original model's behavior while maintaining generation quality and prompt fidelity.

\begin{table*}[t]
  \caption{Results comparing KV Cache policy to base model on MS-COCO 2017 dataset \cite{lin2015microsoftcococommonobjects}.} %
  \centering
  \footnotesize
  \setlength{\tabcolsep}{1pt}
  \renewcommand{\arraystretch}{1.1}
  \begin{tabular}{llcccccccc}
    \toprule
    \multirow{2}{*}{\textbf{Method}} & \multirow{2}{*}{\textbf{Budget}} & \multicolumn{4}{c}{\textbf{Infinity-2B}} & \multicolumn{4}{c}{\textbf{Infinity-8B}} \\
    \cmidrule(lr){3-6} \cmidrule(lr){7-10}
    & & \textbf{KV Cache} & \textbf{FID}$\downarrow$ & \textbf{LPIPS}$\downarrow$ & \textbf{PSNR}$\uparrow$ & \textbf{KV Cache} & \textbf{FID}$\downarrow$ & \textbf{LPIPS}$\downarrow$ & \textbf{PSNR}$\uparrow$ \\
    \midrule
    Full Cache & 100\% & 38550 MB & -- & -- & -- & 84328 MB & -- & -- & -- \\
    \midrule
    StreamingLLM \cite{xiao2024efficient} & 20\% & 7800 MB & 3.52 & 0.139 & 22.00 & 17062 MB & 3.29 & 0.111 & 22.32 \\
    HACK \cite{qin2025headawarekvcachecompression} & 20\% & 7800 MB & 3.29 & 0.122 & 22.79 & 17062 MB & 2.80 & 0.110 & 21.99 \\
    SnapKV \cite{li2024snapkvllmknowslooking} & 20\% & 7800 MB & 3.11 & 0.108 & 23.26 & 17062 MB & 3.09 & 0.102 & 22.72 \\
    ScaleKV \cite{li2025scalekv} & 20\% & 7800 MB & 1.94 & 0.079 & 24.93 & 17062 MB & 1.44 & 0.059 & 25.60 \\
    \rowcolor{gray!15} \textbf{\ourmethod} & 20\% & 7800 MB & \textbf{1.02} & \textbf{0.044} & \textbf{28.43} & 17062 MB & \textbf{1.04} & \textbf{0.043} & \textbf{27.44} \\
    \midrule
    StreamingLLM \cite{xiao2024efficient} & 10\% & 3900 MB & 4.87 & 0.199 & 19.58 & 8531 MB & 4.78 & 0.152 & 20.57 \\
    HACK \cite{qin2025headawarekvcachecompression} & 10\% & 3900 MB & 4.30 & 0.162 & 21.05 & 8531 MB & 3.80 & 0.135 & 20.94 \\
    SnapKV \cite{li2024snapkvllmknowslooking} & 10\% & 3900 MB & 4.42 & 0.148 & 21.56 & 8531 MB & 4.77 & 0.144 & 20.65 \\
    ScaleKV \cite{li2025scalekv} & 10\% & 3900 MB & 2.59 & 0.113 & 22.88 & 8531 MB & 2.13 & 0.085 & 23.25 \\
    \rowcolor{gray!15} \textbf{\ourmethod} & 10\% & 3900 MB & \textbf{1.67} & \textbf{0.073} & \textbf{25.40} & 8531 MB & \textbf{1.80} & \textbf{0.072} & \textbf{24.41} \\
    \midrule
    StreamingLLM \cite{xiao2024efficient} & 4\% & 1590 MB & 6.15 & 0.239 & 18.55 & 3478 MB & 6.57 & 0.197 & 19.11 \\
    HACK \cite{qin2025headawarekvcachecompression} & 4\% & 1590 MB & 5.82 & 0.231 & 18.73 & 3478 MB & 5.57 & 0.185 & 19.34 \\
    SnapKV \cite{li2024snapkvllmknowslooking} & 4\% & 1590 MB & 4.85 & 0.183 & 20.18 & 3478 MB & 6.57 & 0.185 & 19.21 \\
    ScaleKV \cite{li2025scalekv} & 4\% & 1590 MB & 4.92 & 0.168 & 20.88 & 3478 MB & 3.71 & 0.137 & 20.69 \\
    \rowcolor{gray!15} \textbf{\ourmethod} & 4\% & 1590 MB & \textbf{2.70} & \textbf{0.127} & \textbf{22.14} & 3478 MB & \textbf{2.71} & \textbf{0.111} & \textbf{21.78} \\
    \bottomrule
  \end{tabular}
  \label{tab:infinity_budget_results}
\end{table*}

\subsection{Main Results}

\paragraph{Fidelity to Base Model} Table~\ref{tab:infinity_budget_results} reports quantitative fidelity results on MS-COCO 2017. Overall, \ourmethod provides the strongest trade-off between compression and reconstruction quality with images that are hard to distinguish even at very high compression ratio (Figure \ref{fig:eight-image-grid}). In particular, on Infinity-2B, \ourmethod at a 10\% budget outperforms the strongest baseline at a 20\% budget on PSNR, LPIPS, and FID, indicating that it preserves the behavior of the uncompressed model while using only half of the KV-cache budget. This result suggests that \ourmethod allocates cache capacity more effectively than prior approaches under tight memory constraints. In Figure \ref{fig:pareto}, we highlight the relative improvement of HeatKV across additional budgets as well as inference runtime compared to full cache and ScaleKV. Using a flash-attention \cite{daoflashattention} style triton kernel, \ourmethod matches the full cache latency with batch size 2 and 4. The memory advantage of \ourmethod unlocks the possibility for running inference at larger batch sizes, which effectively also enables faster generation speed compared to full-cache inference.

Among the baselines, ScaleKV is the strongest competitor, particularly on Infinity-8B, but its binary per-layer allocation limits how precisely it can distribute the budget. StreamingLLM remains a strong lightweight baseline despite its simple strategy of retaining sink and recent tokens, highlighting the importance of early and late context in visual autoregressive generation. Overall, these comparisons suggest that finer-grained budget allocation is necessary to preserve fidelity under strong compression.


\begin{figure}[b]
\centering
\begin{minipage}[t]{0.31\textwidth}
\centering
\begin{tikzpicture}
\begin{axis}[
    width=6.2cm,
    height=4.2cm,
    ylabel style={yshift=-2pt},
    xlabel style={yshift=2pt},
    xlabel={Cache Budget (\%)},
    ylabel={FID},
    xmin=2.5, xmax=21,
    ymin=0.5, ymax=6.5,
    xtick={4,7,10,15,20},
    grid=major,
    grid style={dotted, gray!40},
    axis lines=left,
    tick style={black},
    tick label style={font=\small},
    label style={font=\small},
    title style={font=\small\bfseries},
    clip=false,
    line width=0.9pt,
    mark size=2pt,
    legend style={
        at={(0.5,1.17)},
        anchor=north,
        draw=none,
        fill=none,
        font=\small
    },
    legend columns=3,
]
\addplot[
    color=softgreen,
    mark=o,
]
coordinates {
    (4,2.7005)
    (7,2.097607150979286)
    (10,1.6679)
    (15,1.1710111670236074)
    (20,1.0225)
};
\addlegendentry{\ourmethod}

\addplot[
    color=softblue,
    mark=triangle*,
]
coordinates {
    (4,4.9156) (10,2.4875) (20,1.9408)
};
\addlegendentry{ScaleKV}

\addplot[
    color=softred,
    mark=diamond*,
]
coordinates {
    (4,5.8150) (10,4.2999) (20,3.2906)
};
\addlegendentry{HACK}

\end{axis}
\end{tikzpicture}
\end{minipage}\hfill%
\begin{minipage}[t]{0.65\textwidth}
\centering
\begin{tikzpicture}
\begin{axis}[
    ybar,
    width=7.8cm,
    height=4.2cm,
    bar width=7pt,
    enlarge x limits=0.12,
    ymin=0,
    ymax=1.6,
    ylabel={Latency (s)},
    xlabel={Batch size},
    ylabel style={yshift=-2pt},
    xlabel style={yshift=2pt},    
    axis lines=left,
    enlarge x limits={
    abs=0.7cm
},
    clip=false,
    symbolic x coords={1,2,4,8,16},
    xtick=data,
    ytick={0,0.5,1,1.5},
    line width=0.9pt,
    mark size=2pt,
    legend style={
        at={(0.5,1.17)},
        anchor=north,
        draw=none,
        fill=white,
        font=\small
    },
    legend columns=3,
    legend image code/.code={
        \draw[#1, draw=none] (0cm,-0.08cm) rectangle (0.18cm,0.08cm);
    },
    nodes near coords,
    every node near coord/.append style={
        font=\scriptsize,
        anchor=south,
        yshift=1pt,
        line width=0.3pt,
        text=black,
    },
    point meta=explicit symbolic
]

\addplot+[fill=softbrown, draw=none] coordinates { %
    (1,0.82) [0.82]
    (2,0.58) [0.58]
    (4,0.45) [0.45]
    (8,0) [OOM]
    (16,0) [OOM]
};

\addplot+[fill=softblue, draw=none] coordinates { %
    (1,1.32) [1.32]
    (2,0.96) [0.96]
    (4,0.79) [0.79]
    (8,0.73) [0.73]
    (16,0.69) [0.69]
};

\addplot+[fill=softgreen, draw=none] coordinates { %
    (1,0.87) [0.87]
    (2,0.58) [0.58]
    (4,0.45) [0.45]
    (8,0.40) [0.40]
    (16,0.38) [0.38]
};

\legend{Full, ScaleKV, \ourmethod}
\end{axis}
\end{tikzpicture}
\end{minipage}\hfill%
\caption{Infinity-2B results on  MS-COCO 2017 and H100 per-image latency using 10\% budget.}
\label{fig:pareto}
\end{figure}

\paragraph{Prompt Alignment and Perceptual Quality} Table \ref{tab:geneval_hps} shows how \ourmethod maintains strong prompt alignment on GenEval down to 4\% budget. We note that prompt alignment is less affected by KV-cache compression than other metrics and therefore the difference between the different methods is subtle. Still, \ourmethod preserves key features and relations even at high compression. On HPSv2.1, perceived image quality using \ourmethod remains high as the budget decreases. Notably, our method at 4\% outperforms all the other methods at 20\% budget on Infinity-2B~\cite{han2025infinity}. While ScaleKV performs well overall, it degrades at low budgets, and SnapKV works well on 2B but less so on 8B. Overall, methods with uneven budget allocation across layers or heads perform better on 8B, indicating that allocation becomes more critical as model size grows.

\paragraph{Generation speed}
Figure \ref{fig:pareto} (right) shows the generation speed using different batch sizes and cache policies. At batch size 1 there is a 6\% overhead and the workload is launch-bound: roughly half of the runtime is host-side overhead for both methods, with the GPU largely idle. HeatKV's additional cache management contributes a small fixed cost in exactly this regime, while the savings from compression, which affect only the cache-dependent part of the computation, are at their smallest. Because this cost is constant in batch size, it amortizes as the batch grows. The practical benefit of HeatKV is compression of memory which increases the possible batch size. Comparing throughputs (i.e. the inverse of per-image latency) from Fig \ref{fig:pareto}, HeatKV's best achievable throughput exceeds the baseline's by around 18\%, which is memory-capped. 

\subsection{Ablation Study}
\begin{table}[t]
  \begin{minipage}[t]{0.60\textwidth}
  \centering
  \small
  \caption{Results on Infinity-2B and Infinity-8B using GenEval \cite{geneval} and HPSv2.1 \cite{wu2023human}}
    \vspace{1.2em}
  \setlength{\tabcolsep}{1.5pt}
  \renewcommand{\arraystretch}{1.1}
  \begin{tabular}{llcccc}
    \toprule
    \multirow{2}{*}{\textbf{Method}} & \multirow{2}{*}{\textbf{Budget}} & \multicolumn{2}{c}{\textbf{Infinity-2B}} & \multicolumn{2}{c}{\textbf{Infinity-8B}} \\
     \cmidrule(lr){3-4} \cmidrule(lr){5-6}
    & & \textbf{GenEval} & \textbf{HPSv2.1} & \textbf{GenEval} & \textbf{HPSv2.1} \\
    \midrule
    Full Cache &\multirow{1}{*}{100\%} & 0.680 & 30.50 & 0.826 & 30.90 \\
    \midrule
    StreamingLLM & 20 \% & 0.679 & 30.36 & 0.814 & 30.40  \\
    HACK & 20 \% & 0.677 & 30.17 & 0.820 & 30.56 \\
    SnapKV & 20 \% & 0.677 & 30.31 & 0.815 & 30.51 \\
    ScaleKV & 20 \% & 0.678 & 30.42 & 0.821 & 30.81 \\
    \rowcolor{gray!15} \textbf{\ourmethod} & 20 \% & \textbf{0.684} & \textbf{30.51} & \textbf{0.824} & \textbf{30.86} \\
    \midrule
    StreamingLLM & 10 \% & 0.677 & 30.21 & 0.816 & 30.26  \\
    HACK & 10 \% & 0.675 & 29.99 & 0.813 & 30.40 \\
    SnapKV & 10 \%  & 0.675 & 30.05 & 0.813 & 30.18 \\
    ScaleKV & 10 \%  & 0.675 & 30.34 & \textbf{0.819} & 30.73 \\
    \rowcolor{gray!15} \textbf{\ourmethod} & 10 \% & \textbf{0.684} & \textbf{30.50} & 0.816 & \textbf{30.75} \\
    \midrule
    StreamingLLM & 4\% & 0.665 & 29.77 & 0.819 & 30.04 \\
    HACK & 4\% & 0.665 & 29.59 & 0.814 & 30.07 \\
    SnapKV & 4\% & 0.682 & 30.04 & 0.814 & 29.86 \\
    ScaleKV & 4\% & 0.669 & 29.74 & 0.820 & 30.35 \\
    \rowcolor{gray!15} \textbf{\ourmethod} & 4\% & \textbf{0.683} & \textbf{30.45} & \textbf{0.821} & \textbf{30.66} \\
    \bottomrule
  \end{tabular}
  \label{tab:geneval_hps}
  \end{minipage}
  \hfill
   \begin{minipage}[t]{0.38\textwidth}
      \footnotesize
      \centering
      \caption{Results on MS-COCO (10\% budget) with different: \\
      a) number of calibration prompts $|\mathcal{D}|$}
      \setlength{\tabcolsep}{5pt}
      \renewcommand{\arraystretch}{1.1}
      \begin{tabular}{lccc}
        \toprule
        $|\mathcal{D}|$ & \textbf{FID}$\downarrow$ & \textbf{LPIPS}$\downarrow$ & \textbf{PSNR}$\uparrow$ \\
        \midrule
        1 & 1.67 & 0.0734 & 25.38 \\
        2 & 1.67 & 0.0735 & 25.37 \\
        5 & 1.66 & 0.0732 & 25.40 \\
        10 & 1.67 & 0.0732 & 25.40 \\
        \bottomrule
      \end{tabular}
      \label{tab:ablation}
      \centering
      \caption*{b) number of sink scales $s$}
      \setlength{\tabcolsep}{5pt}
      \renewcommand{\arraystretch}{1.1}
      \begin{tabular}{lccc}
        \toprule
        \textbf{$s$}& \textbf{FID}$\downarrow$ & \textbf{LPIPS}$\downarrow$ & \textbf{PSNR}$\uparrow$ \\
        \midrule
        1 & 1.71 & 0.075 & 25.26 \\
        2 & 1.67 & 0.074 & 25.37 \\
        3 & 1.67 & 0.073 & 25.40 \\
        4 & 1.69 & 0.074 & 25.32 \\
        5 & 1.71 & 0.075 & 25.25 \\
        6 & 1.79 & 0.079 & 25.00 \\
        \bottomrule
      \end{tabular}
        \centering
        \caption*{c) pruning methods}
        \renewcommand{\arraystretch}{1.15}
        \begin{tabular}{lccc}
        \hline
        \textbf{Method} & \textbf{FID}$\downarrow$ & \textbf{LPIPS}$\downarrow$ & \textbf{PSNR}$\uparrow$ \\
        \hline
        Naive & 2.21 & 0.101 & 23.50 \\
        B-HeatKV & 1.78 & 0.078 & 25.06 \\
        \ourmethod  & 1.67 & 0.073 & 25.40 \\
        \hline
        \end{tabular}
    \end{minipage}
\end{table}

\paragraph{Calibration Prompts} The S-CAS scores used to rank heads can be computed from a single prompt or averaged across multiple prompts. To test the sensitivity with respect to the calibration-set size, we vary the number of prompts used to compute S-CAS, as shown in Table~\ref{tab:ablation}a). The results support the observations of \cite{li2025scalekv, qin2025headawarekvcachecompression}: attention patterns are highly stable across prompts. Even when increasing the calibration set from 1 prompt to 10 prompts, all evaluation metrics remain within a narrow range. 

\paragraph{Number of Sink Scales} Table~\ref{tab:ablation}b) studies the effect of the number of sink scales on Infinity-2B at a 10\% budget. Performance remains largely stable up to four sink scales, after which it begins to degrade. This suggests that preserving the earliest scales is beneficial, but only up to the point where the additional sink tokens start displacing too much useful capacity elsewhere in the cache. Because the earliest scales contain relatively few tokens, increasing the sink from two to four scales has only a modest budget cost after which the fifth scale more than doubles the size of the sink. %

\paragraph{Pruning Algorithm} In order to demonstrate the importance of the different components in our method, we have conducted an ablation study of the budget allocation strategy and pruning method. Results are shown in Table~\ref{tab:ablation}c). In its simplest form, we use a binary budget allocation for each head (retain or discard all tokens) with the naive method ($E_k = G_k \setminus G_{k-1}$). By using Binary-\ourmethod, the results improve noticeably and it outperforms the other methods in Table~\ref{tab:infinity_budget_results}. By relaxing the binary budget constraint and allowing each head to retain or discard tokens from different scales, our \ourmethod policy further improves the results. 

\paragraph{Calibration Data Domain Shift} In order to verify that the score calculation is robust to domain shift in the calibration data, we calculate the S-CAS score with a set of 10 prompts from three distinct categories as well as the COCO2017 dataset. Each calibration is tested on each of the distribution-shifted datasets. All experiments are conducted using Infinity-2B at 10\% budget and the results are shown in Table \ref{tab:combined_metrics} . We note that HeatKV is robust to domain shift, since calibrating on one domain and testing on another has very little impact on the final results. These ablations indicate that the head ranking induced by S-CAS is robust and that the distribution of the calibration set does not play an important role for generation quality.

\paragraph{Pruning Methods} Table \ref{tab:token_counts} shows the maximum number of tokens in the cache for each scale. It clearly shows how the soft budget method over-allocates (highlighted in red) when it moves allocation from later layers to earlier ones. The naive allocation significantly under-allocates for all but the last scale as it conservatively starts using the next scales token mask as the current scales starts to avoid the temporary overvaluations. That is why the last scale is so close to the budget as it can drop the cache after each layer to ensure no budget breaches. In Table \ref{tab:pruning_results} we include an ablation on the greedy pruning algorithm. It clearly shows how the greedy pruning strategy is only marginally beaten by the soft budget limit.

\begin{table*}[t]
\footnotesize
\centering
\caption{FID, LPIPS, and PSNR across different calibration and test set combinations for HeatKV, the last row shows ScaleKV results.}
\label{tab:combined_metrics}
\begin{tabular}{lcccccccccccc}
\toprule
& \multicolumn{4}{c}{FID $\downarrow$}
& \multicolumn{4}{c}{LPIPS $\downarrow$}
& \multicolumn{4}{c}{PSNR $\uparrow$} \\
\cmidrule(lr){2-5} \cmidrule(lr){6-9} \cmidrule(lr){10-13}
Cal.
& Car & Port & Land & Coco
& Car & Port & Land & Coco
& Car & Port & Land & Coco \\
\midrule
Car
& 0.944 & 0.987 & 2.486 & 1.699
& 0.083 & 0.075 & 0.101 & 0.074
& 21.61 & 29.54 & 25.13 & 25.31 \\

Port
& 0.950 & 0.966 & 2.464 & 1.677
& 0.082 & 0.074 & 0.100 & 0.074
& 21.46 & 29.62 & 25.15 & 25.34 \\

Lands
& 0.943 & 0.986 & 2.475 & 1.684
& 0.083 & 0.075 & 0.100 & 0.074
& 21.31 & 29.54 & 25.16 & 25.33 \\

Coco
& 0.934 & 0.906 & 2.459 & 1.67
& 0.082 & 0.073 & 0.099 & 0.073
& 21.68 & 29.69 & 25.42 & 25.40 \\
\midrule
S-KV
& 1.676 & 2.109 & 3.661 & 2.59
& 0.125 & 0.120 & 0.145 & 0.113
& 19.29 & 26.86 & 23.41 & 22.88 \\
\bottomrule
\end{tabular}

\vspace{2pt}
\parbox{\textwidth}{\scriptsize
\textit{Note:} Cal. = calibration set;
Port = Portrait; Land = Landscape; Coco = Coco2017; S-KV = ScaleKV.
}
\end{table*}

\begin{table}[t]
\centering
\footnotesize
\begin{minipage}[t]{0.56\columnwidth}
\centering
\caption{Token counts at different scales using Infinity-2B with a 10\% budget.}
\label{tab:token_counts}
\begin{tabular}{cccccc}
\toprule
Scale & Tokens & Naive & Greedy & Soft & Budget \\
\midrule
7  & 256  & 187,252 & 266,752 & 266,752 & 328,960 \\
8  & 400  & 204,492 & 328,452 & 328,452 & 328,960 \\
9  & 576  & 217,704 & 328,092 & \cellcolor{lightred}346,716 & 328,960 \\
10 & 1024 & 209,288 & 328,252 & \cellcolor{lightred}347,240 & 328,960 \\
11 & 1600 & 217,984 & 326,452 & \cellcolor{lightred}341,928 & 328,960 \\
12 & 2304 & 324,548 & 324,548 & \cellcolor{lightred}360,768 & 328,960 \\
\bottomrule
\end{tabular}
\end{minipage}
\hfill
\begin{minipage}[t]{0.42\columnwidth}
\centering
\caption{Comparison of pruning methods using Infinity-2B model on MS-COCO.}
\label{tab:pruning_results}
\begin{tabular}{lccc}
\toprule
Pruning & FID & LPIPS & PSNR \\
\midrule
Naive & 2.21 & 0.101 & 23.50 \\
Greedy & 1.67 & 0.073 & 25.40 \\
Soft & 1.65 & 0.072 & 25.50 \\
\bottomrule
\end{tabular}
\end{minipage}
\end{table}

\section{Conclusions and Future Work}

In this work, we have proposed a novel method for KV-cache compression in VAR models and demonstrated strong results across all benchmarks and models. As a consequence, it potentially allows VAR models to be used on consumer hardware where memory capacity is limited. Alternatively, our method can be used to increase the capacity of an existing model by serving more concurrent users simultaneously. However, we have not explored other ways of optimizing VAR models that are complementary to KV-cache compression. This includes speeding up the generation process by exploiting sparsity in the attention mechanism \cite{guo2025fastvar, li2026sparvar}, as well as model quantization \cite{xie2024litevar, liu2026ptqarvg}. We note that these optimizations can be combined with our proposed KV-cache compression algorithm and are left for future work to explore this research direction.

\begin{ack}
We thank Danny Loh for supporting this work and, Eva Li for her feedback and comments. Finally, we thank Kunjun Li for sharing the code for ScaleKV with the research community. P. Giselsson acknowledges support from the ELLIIT Strategic Research Area, the Wallenberg AI, Autonomous Systems, and Software Program (WASP) funded by the Knut and Alice Wallenberg Foundation, and the Swedish Research Council.
\end{ack}

\bibliographystyle{ieeetr}
{\small
\bibliography{bib}
}

\newpage
\appendix

\section{Additional algorithms}
\label{appendix:algorithms}

\begin{algorithm}[!h]
\caption{\textsc{CacheSizeAfterLayer}$(k,\ell,G_k,G_{k-1},E_k)$}
\label{alg:cache-size-after-layer}
\begin{algorithmic}
\Require scale $k$, current layer $\ell$, target pruning set $G_k$, previous pruning set $G_{k-1}$, early pruning set $E_k$
\State $A_k \gets G_{k-1} \cup E_k$
\State $M \gets 0$

\For{each layer $\ell' = 1$ to $L$}
    \For{each head $h = 1$ to $H$}
        \If{$\ell' \le \ell$}\Comment{layers already executed have applied all scale-$k$ removals}
        
            \State $M \gets M + \textsc{CachedTokens}(k,\ell',h,G_k)$ 
        \Else    \Comment{future layers only include removals done before scale $k$}
            \State $M \gets M + \textsc{CachedTokens}(k,\ell',h,A_k)$ 
            
        \EndIf

    \EndFor
\EndFor

\State \Return $M$
\end{algorithmic}
\end{algorithm}

For the binary case \textsc{CacheSizeAfterLayer} uses

\[
\textsc{\colorbox{blue!30}{CachedTokens}}(k,\ell,h,P)
=
\begin{cases}
c_s, & (\ell,h) \in P,\\
c_k, & \text{otherwise}.
\end{cases}
\]

But for the scale dependent case the cache consist of head-scales, a subset of all previous scales.
\[
\textsc{\colorbox{green!30}{CachedTokens}}(k,\ell,h,P)
=
c_s
+
\sum_{i=s+1}^{k}
t_i \cdot
\mathbf{1}\!\left[(i,\ell,h)\notin P\right].
\]

\section{Optimality of Binary Greedy Early Pruning}
\label{appendix:proof}
Let \(\mathcal{E}_k\) be any subset of \(G_k \setminus G_{k-1}\), an early-pruning set such that the cache usage at scale \(k\) and layer \(\ell\) does not exceed a budget \(B\) for all \(\ell\).
\begin{equation}
    \mathcal{E}_k
\in \left\{
E \subseteq G_k \setminus G_{k-1}
\;\middle|\;
f_{\mathrm{past}}^{(k)}(\ell)
+ f_{\mathrm{future}}^{(k)}(\ell)
- f_{\mathrm{early}}^{(k)}(\ell;E)
\le B
\;\;\forall \ell
\right\}.
\end{equation}

We define the optimal early-pruning set at scale \(k\) as
\begin{equation}
    E_k^{(\mathrm{opt})}
    = \arg\min_{\mathcal{E}_k \subseteq G_k \setminus G_{k-1}} |\mathcal{E}_k|
    \quad \text{s.t.} \quad
    f_{\mathrm{cache}}^{(k)}(\ell;\mathcal{E}_k) \le B
    \;\; \forall \ell,
    \label{eq:binary_optim}
\end{equation}
where \(G_k \setminus G_{k-1}\) is the set of heads eligible for early pruning, and
\begin{equation}
    f_{\mathrm{cache}}^{(k)}(\ell;\mathcal{E}_k)
    =
    f_{\mathrm{past}}^{(k)}(\ell)
    +
    f_{\mathrm{future}}^{(k)}(\ell)
    -
    f_{\mathrm{early}}^{(k)}(\ell;\mathcal{E}_k).
\end{equation}

Here,
\(f_{\mathrm{past}}^{(k)}(\ell)\) is the number of tokens already committed in layers up to \(\ell\),
\(f_{\mathrm{future}}^{(k)}(\ell)\) is the number of tokens that would be cached in layers after \(\ell\) without additional early pruning,
and \(f_{\mathrm{early}}^{(k)}(\ell;\mathcal{E}_k)\) is the reduction due to the heads in \(\mathcal{E}_k\) being pruned early.

\medskip
\noindent
For a head \(h\), let $r_\ell(h)$ denote its contribution to \(f_{\mathrm{early}}^{(k)}(\ell;\mathcal{E}_k)\). Since each early-pruned head removes the same number of tokens per affected future layer, \(r_\ell(h)\) depends only on the layer of \(h\). In particular, if
\[
    \mathrm{layer}(h_1) \ge \mathrm{layer}(h_2),
\]
then for every \(\ell\),
\begin{equation}
    r_\ell(h_1) \ge r_\ell(h_2).
    \label{eq:dominance}
\end{equation}
That is, pruning a head from a deeper layer yields at least as much cache reduction as pruning a head from a shallower layer, at every cache checkpoint \(\ell\).

\medskip
\noindent
Let \(E_k^{(\mathrm{greedy})}\) be the set obtained by greedily selecting heads in decreasing order of layer index, i.e., from the deepest layer to the shallowest, until the constraint
\[
    f_{\mathrm{cache}}^{(k)}(\ell;\mathcal{E}_k) \le B
    \qquad \forall \ell
\]
is satisfied.

We now show that this greedy solution is optimal. Let \(E_k^{(\mathrm{opt})}\) be any optimal feasible set. Suppose \(E_k^{(\mathrm{opt})}\) is not equal to the greedy set. Then there exists a head \(h \in E_k^{(\mathrm{opt})}\) and a head \(g \notin E_k^{(\mathrm{opt})}\) such that
\[
    \mathrm{layer}(g) \ge \mathrm{layer}(h),
\]
because the greedy solution always prefers deeper heads first. Consider the exchanged set
\[
    E_k' = (E_k^{(\mathrm{opt})}\setminus\{h\}) \cup \{g\}.
\]
By \eqref{eq:dominance}, for every \(\ell\),
\[
    f_{\mathrm{early}}^{(k)}(\ell;E_k')
    \ge
    f_{\mathrm{early}}^{(k)}(\ell;E_k^{(\mathrm{opt})}),
\]
and therefore
\[
    f_{\mathrm{cache}}^{(k)}(\ell;E_k')
    \le
    f_{\mathrm{cache}}^{(k)}(\ell;E_k^{(\mathrm{opt})})
    \le B.
\]
Hence \(E_k'\) is also feasible, and moreover
\[
    |E_k'| = |E_k^{(\mathrm{opt})}|.
\]

\medskip
\noindent
By repeatedly applying this exchange, any optimal solution can be transformed into the greedy solution without increasing its cardinality and without violating feasibility. Therefore,
\[
    |E_k^{(\mathrm{greedy})}| = |E_k^{(\mathrm{opt})}|,
\]
which proves that the greedy strategy is optimal.

\section{S-CAS stability}
\label{appendix:s-cas-stability}
\begin{figure}[h]
\centering
\begin{tikzpicture}
\begin{axis}[
    width=0.75\linewidth,
    height=0.45\linewidth,
    xlabel={Source scale (1-indexed)},
    ylabel={Rank-position standard deviation},
    xmin=3.5, xmax=12.5,
    ymin=0, ymax=25,
    xtick={4,5,6,7,8,9,10,11,12},
    ymajorgrids,
    xmajorgrids,
    grid style={dashed, gray!25},
    legend pos=north east,
    legend style={draw=none, fill=none},
    mark size=2.2pt,
]

\addplot[
    thick,
    color=green!60!black,
    mark=*,
]
coordinates {
    (4, 7.4264)
    (5, 8.5009)
    (6, 8.1839)
    (7, 7.5749)
    (8, 7.1176)
    (9, 7.0987)
    (10, 6.4509)
    (11, 6.5241)
    (12, 5.1439)
};
\addlegendentry{Infinity-2B}

\addplot[
    thick,
    color=red!60!black,
    mark=square*,
]
coordinates {
    (4, 23.5342)
    (5, 19.5695)
    (6, 18.5146)
    (7, 17.9498)
    (8, 18.8015)
    (9, 17.8053)
    (10, 17.9812)
    (11, 17.3550)
    (12, 16.8709)
};
\addlegendentry{Infinity-8B}

\end{axis}
\end{tikzpicture}
\caption{
Stability of S-CAS head-scale rankings across calibration runs for Infinity-2B and Infinity-8B.
The reported source-scale indices are converted from zero-indexed implementation scales to one-indexed paper scales.
For each scale, we compute the standard deviation of head rank positions across runs after centering by the head-specific mean rank.
}
\label{fig:scas_rank_stability}
\end{figure}


To test whether the head-scale rankings used by HeatKV are stable across calibration samples, we repeat the S-CAS ordering procedure over 10 independently generated calibration sets for both Infinity-2B and Infinity-8B. For each source scale, we record the rank position of every attention head in each run and compute the standard deviation of its rank around the head-specific mean rank. We then average this residual dispersion across all heads. Low dispersion indicates that the same heads are consistently assigned similar pruning priority across calibration runs.

Figure~\ref{fig:scas_rank_stability} reports the resulting rank-position standard deviation for each source scale. The observed rank dispersion is substantially below the total number of heads, indicating that S-CAS induces a stable head-scale ordering rather than one dominated by calibration noise. 

Even in the worst case (source scale 4 for Infinity-8B), the observed standard deviation corresponds to less than 2\% of the total ordering range, indicating that heads move only minimally across calibration runs. This supports the use of a small offline calibration set for constructing the static HeatKV pruning schedule. This analysis complements the metric-level calibration ablation in Table~\ref{tab:ablation}a, which shows that downstream fidelity remains nearly unchanged when varying the number of calibration prompts.

\section{Limitations}
While this paper and previous research have found the attention patterns to be stable in VAR models, \ourmethod is still limited by the stability of these attention patterns. The attention patterns are likely to remain stable as local tokens contain much of the information about what the local patch should look like. In addition, \ourmethod makes the assumption that all tokens within a scale are of the same importance. If that assumption is broken, then it will result in a higher ratio of low importance tokens retained in the cache. This will decrease the efficiency at which we cache the tokens leading to a reduction in performance.

\section{Impact}
This paper improves on KV cache compression for VARs, particularly allowing larger models to run on accelerators with limited memory capacity, which opens up the possibility for smaller entities to generate higher quality images. This could indirectly affect negative aspects of image generation. However, \ourmethod has a higher fidelity to the base model than previous methods. On a VAR model trained with guardrails against malicious image generation, \ourmethod will align closer to the image generated by the original safeguarded model. Thereby limiting a potential user's ability to circumvent these guardrails. Our method also has a positive impact on energy consumption by allowing image generation providers to have fewer accelerators for their service. 

\clearpage
\newpage
\section{Additional Results on GenEval}
Table \ref{tab:geneval_full} shows the overall result on the GenEval dataset, as well as separate component metrics: Two Object evaluates if two objects described in the prompt both appear in the generated image, Position checks if they are arranged in the correct spatial relationship, and Color Attribute checks if the right colors are assigned to the correct objects.

\begin{table*}[h]
  \centering
  \small
  \caption{Results on Infinity-2B and Infinity-8B using GenEval \cite{geneval}.}
  \setlength{\tabcolsep}{1.5pt}
  \renewcommand{\arraystretch}{1.1}
  \begin{tabular}{llcccccccc}
    \toprule
    \multirow{2}{*}{\textbf{Method}} & \multirow{2}{*}{\textbf{Budget}} & \multicolumn{4}{c}{\textbf{Infinity-2B}} & \multicolumn{4}{c}{\textbf{Infinity-8B}} \\
    \cmidrule(lr){3-6} \cmidrule(lr){7-10}
    & & \textbf{Two Obj} & \textbf{Position} & \textbf{Color Attri.} & \textbf{Overall} & \textbf{Two Obj} & \textbf{Position} & \textbf{Color Attri.} & \textbf{Overall} \\
    \midrule
    Full Cache &\multirow{1}{*}{100\%} & 0.77 & 0.28 & 0.59 & 0.680 & 0.96 & 0.57 & 0.77 & 0.826 \\
    \midrule
    StreamingLLM & 20 \% & 0.79 & 0.28 & 0.58 & 0.679 & 0.95 & 0.57 & 0.74 & 0.814 \\
    HACK & 20 \% & 0.78 & 0.27 & 0.57 & 0.677 & 0.96 & 0.57 & 0.75 & 0.820 \\
    SnapKV & 20 \% & 0.77 & 0.27 & 0.58 & 0.677 & 0.95 & 0.57 & 0.73 & 0.815 \\
    ScaleKV & 20 \% & 0.78 & 0.27 & 0.58 & 0.678 & 0.96 & 0.57 & 0.75 & 0.821 \\
    \rowcolor{gray!15} \textbf{\ourmethod} & 20 \% & 0.78 & 0.27 & 0.60 & 0.684 & 0.96 & 0.57 & 0.75 & 0.824 \\
    \midrule
    StreamingLLM & 10 \% &  0.77 & 0.29 & 0.55 & 0.677 & 0.95 & 0.56 & 0.74 & 0.816 \\
    HACK & 10 \% & 0.77 & 0.27 & 0.57 & 0.675 & 0.96 & 0.57 & 0.72 & 0.813 \\
    SnapKV & 10 \%  & 0.77 & 0.27 & 0.58 & 0.675 & 0.95 & 0.56 & 0.72 & 0.813 \\
    ScaleKV & 10 \%  & 0.77 & 0.27 & 0.58 & 0.675 & 0.96 & 0.56 & 0.74 & 0.819 \\
    \rowcolor{gray!15} \textbf{\ourmethod} & 10 \% & 0.78 & 0.28 & 0.60 & 0.684 & 0.96 & 0.55 & 0.74 & 0.816 \\
    \midrule
    StreamingLLM & 4\% & 0.75 & 0.27 & 0.54 & 0.665 & 0.95 & 0.56 & 0.76 & 0.819 \\
    HACK & 4\% & 0.78 & 0.27 & 0.57 & 0.665 & 0.95 & 0.56 & 0.73 & 0.814 \\
    SnapKV & 4\% & 0.78 & 0.26 & 0.60 & 0.682 & 0.94 & 0.54 & 0.75 & 0.814 \\
    ScaleKV & 4\% & 0.77 & 0.26 & 0.55 & 0.669 & 0.96 & 0.56 & 0.73 & 0.820 \\
    \rowcolor{gray!15} \textbf{\ourmethod} & 4\% & 0.78 & 0.28 & 0.59 & 0.683 & 0.95 & 0.57 & 0.75 & 0.821 \\
    \bottomrule
  \end{tabular}
  \label{tab:geneval_full}
\end{table*}

\clearpage
\newpage
\section{Additional Visualizations of Generated Images}

\begin{figure}[!ht]
    \centering
    \setlength{\tabcolsep}{1pt}
    \renewcommand{\arraystretch}{0}
    \begin{tabular}{ccc}
        \includegraphics[width=0.328\linewidth]{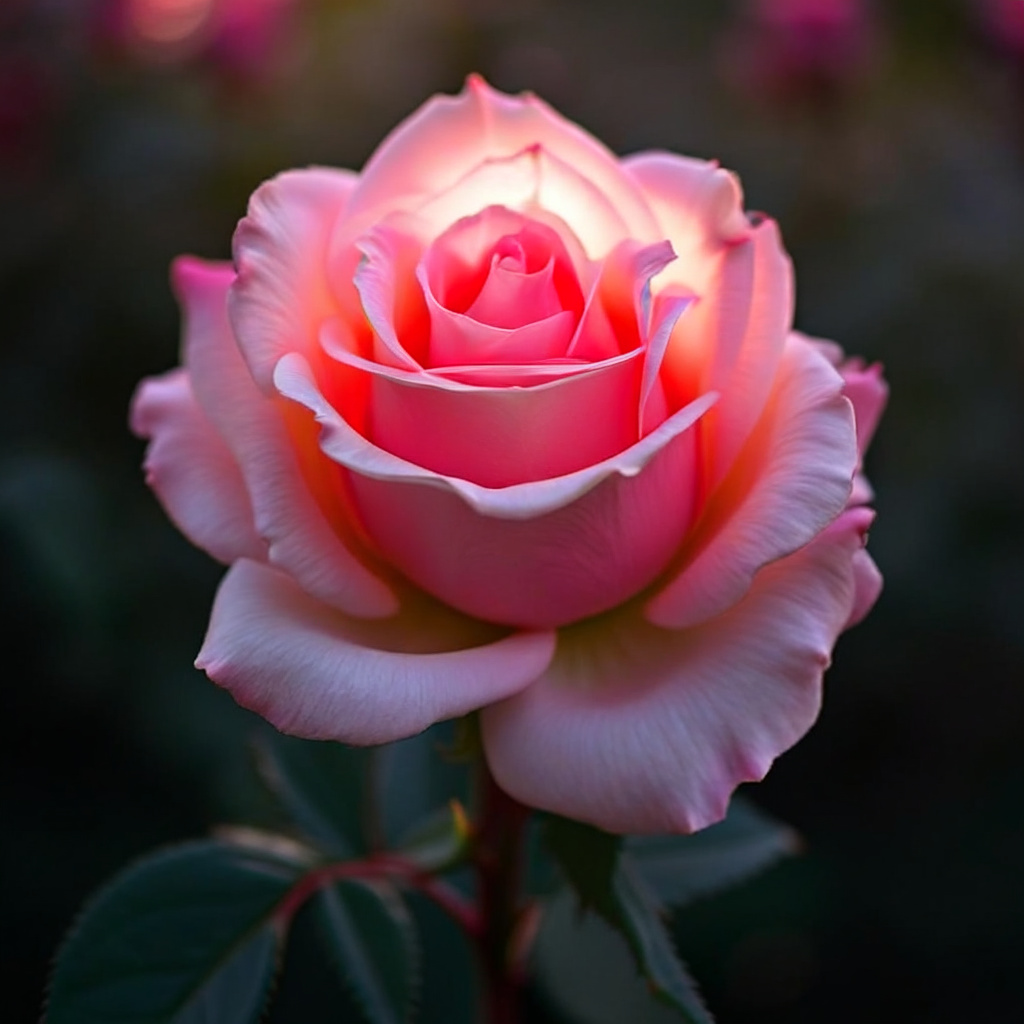} &
        \includegraphics[width=0.328\linewidth]{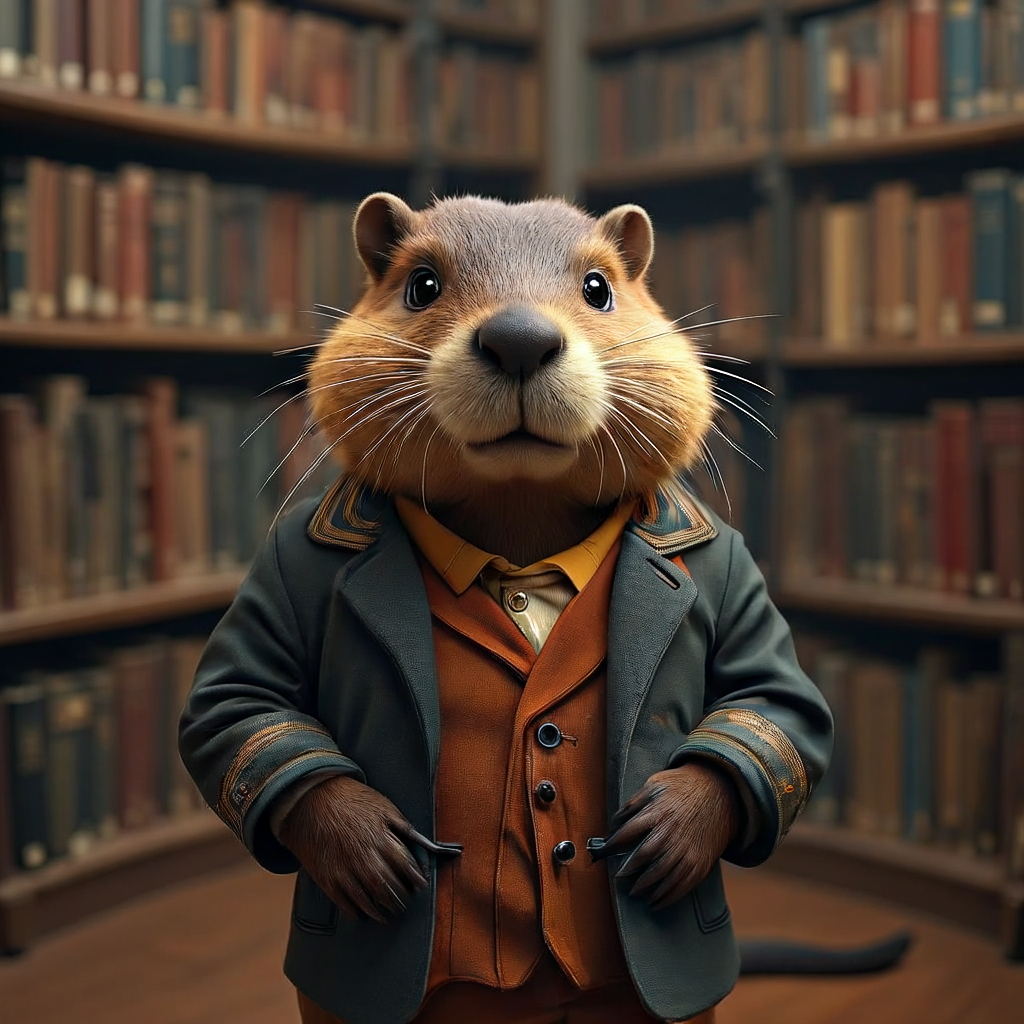} &
        \includegraphics[width=0.328\linewidth]{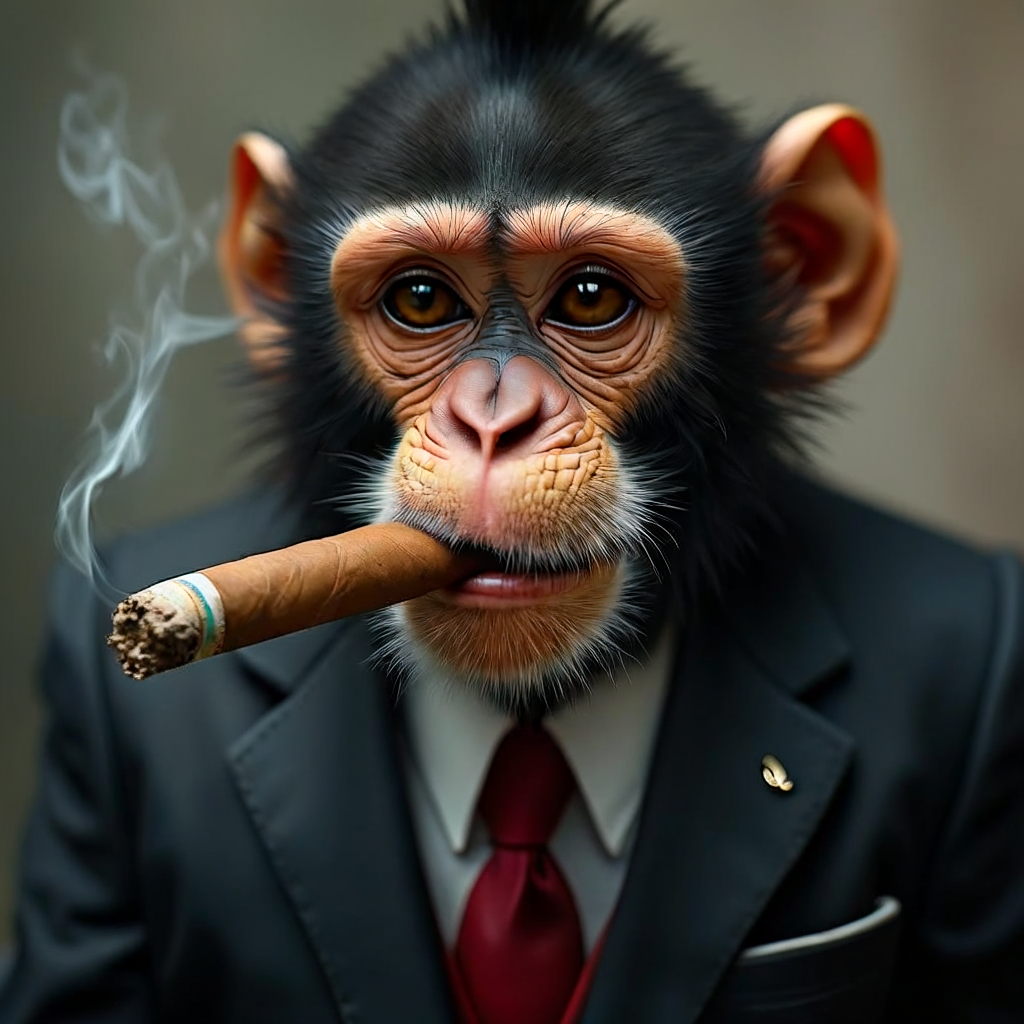} \\
        \includegraphics[width=0.328\linewidth]{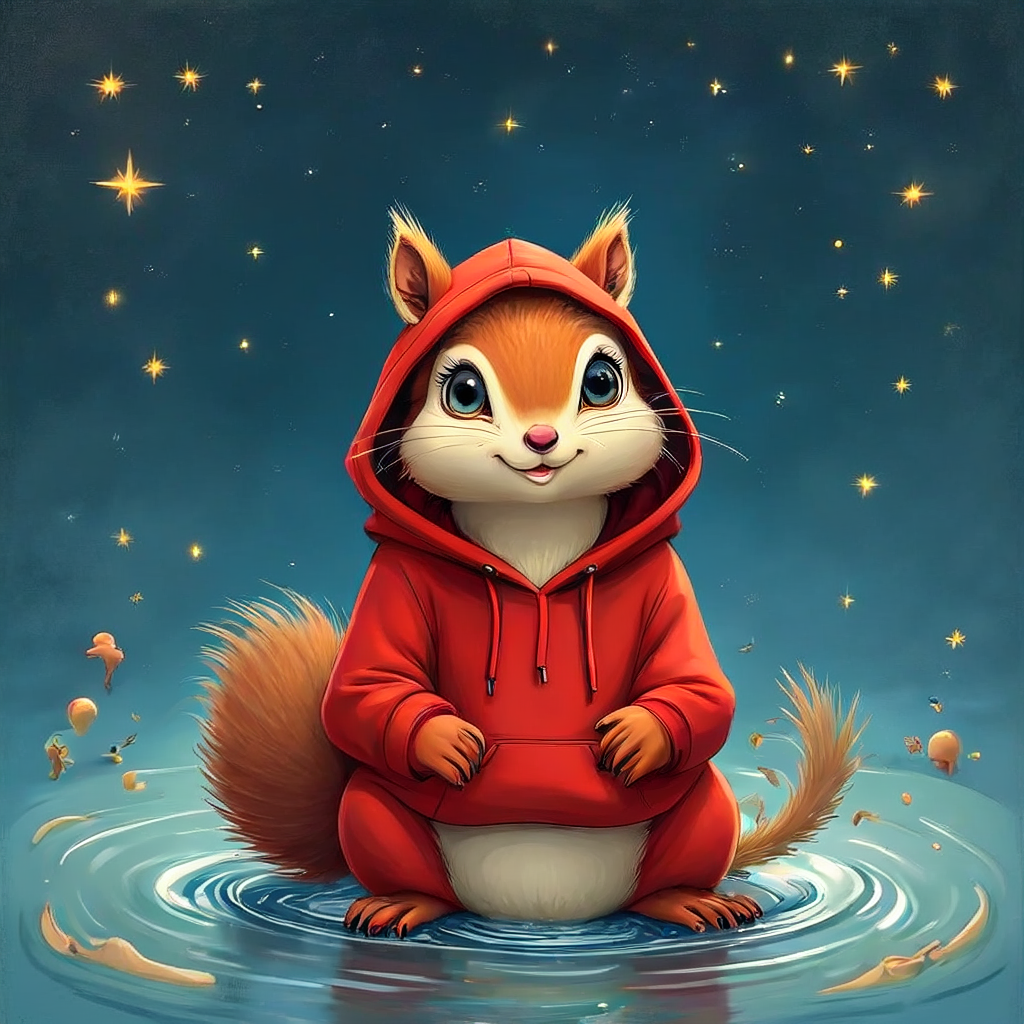} &
        \includegraphics[width=0.328\linewidth]{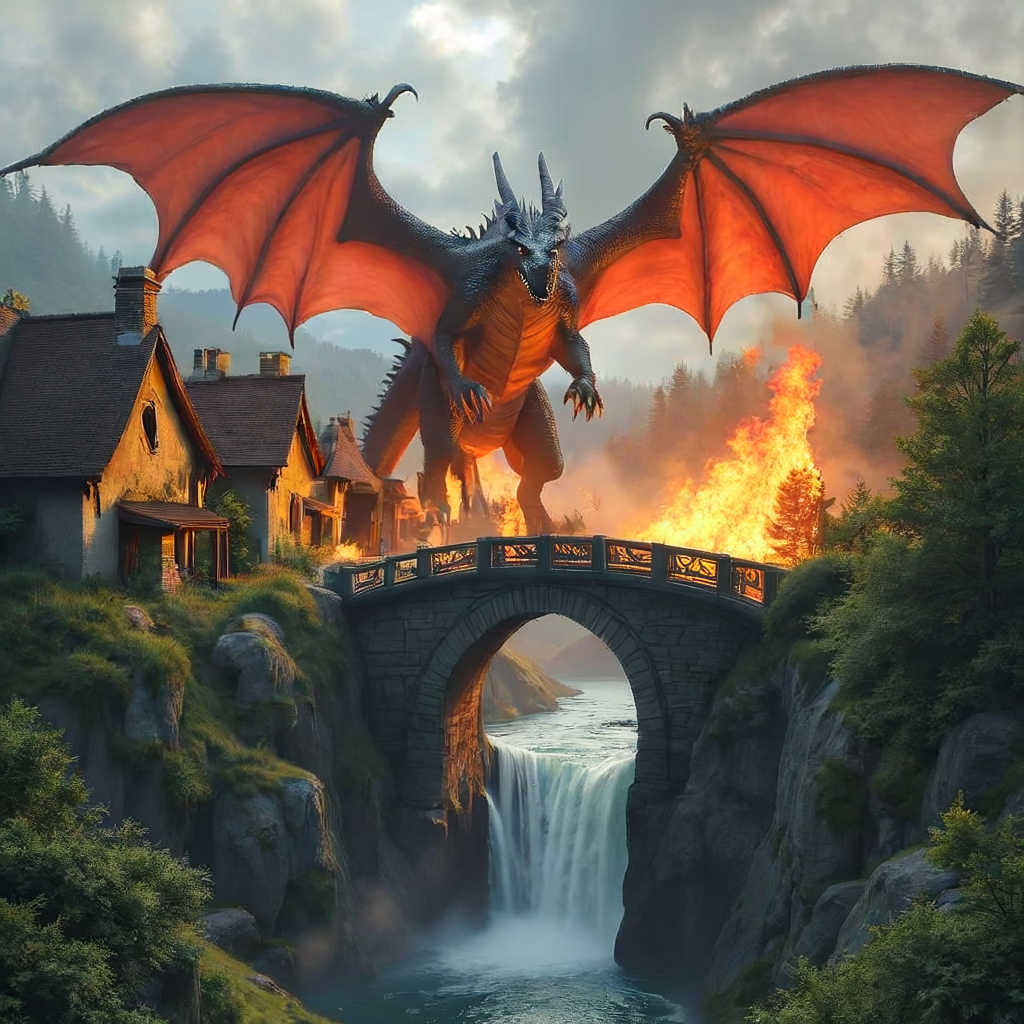} &
        \includegraphics[width=0.328\linewidth]{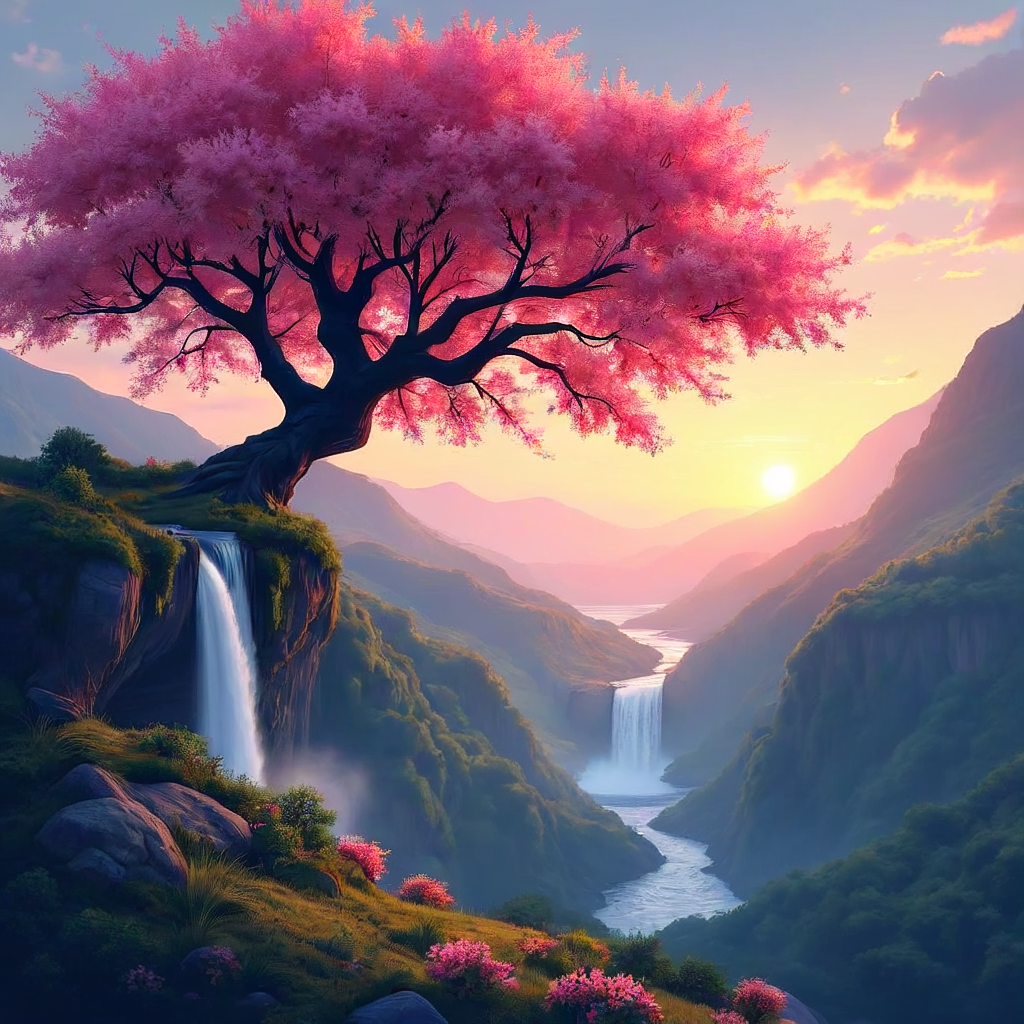} \\
        \includegraphics[width=0.328\linewidth]{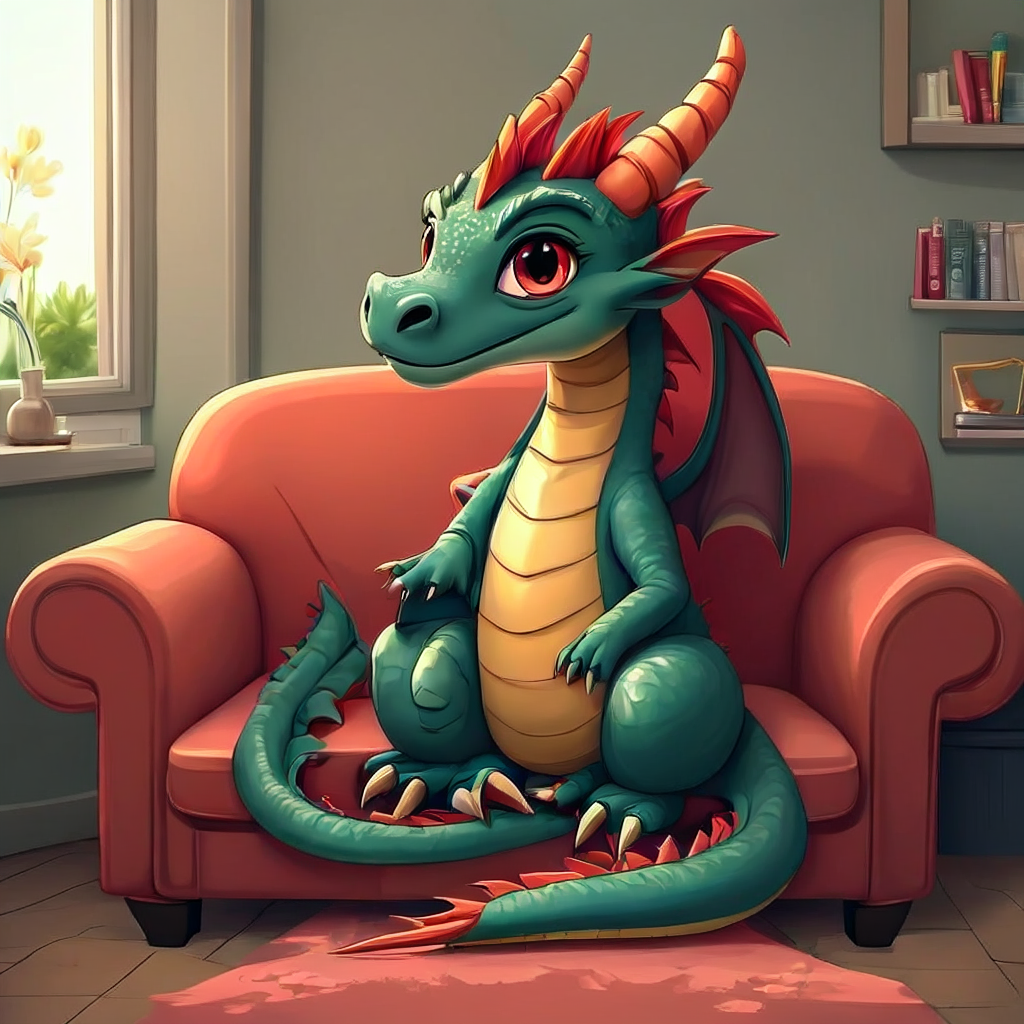} &
        \includegraphics[width=0.328\linewidth]{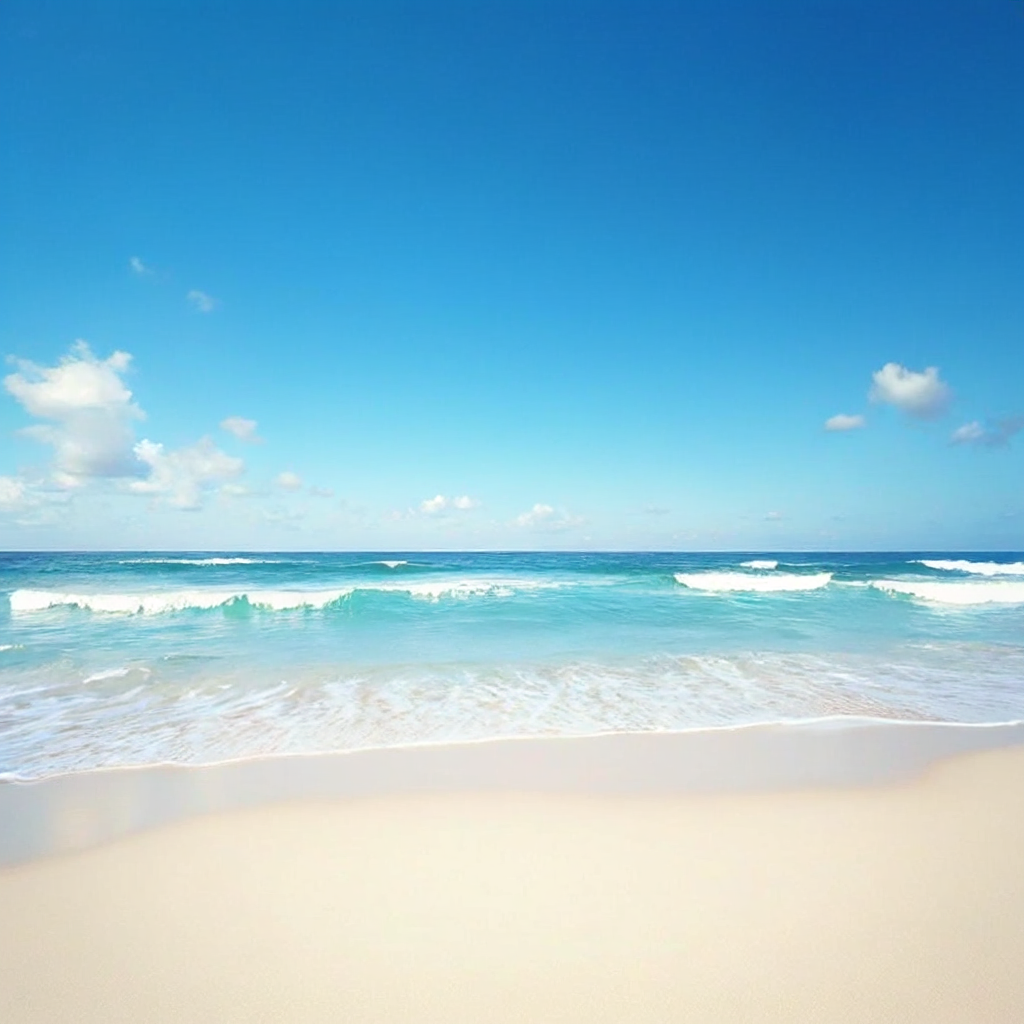} &
        \includegraphics[width=0.328\linewidth]{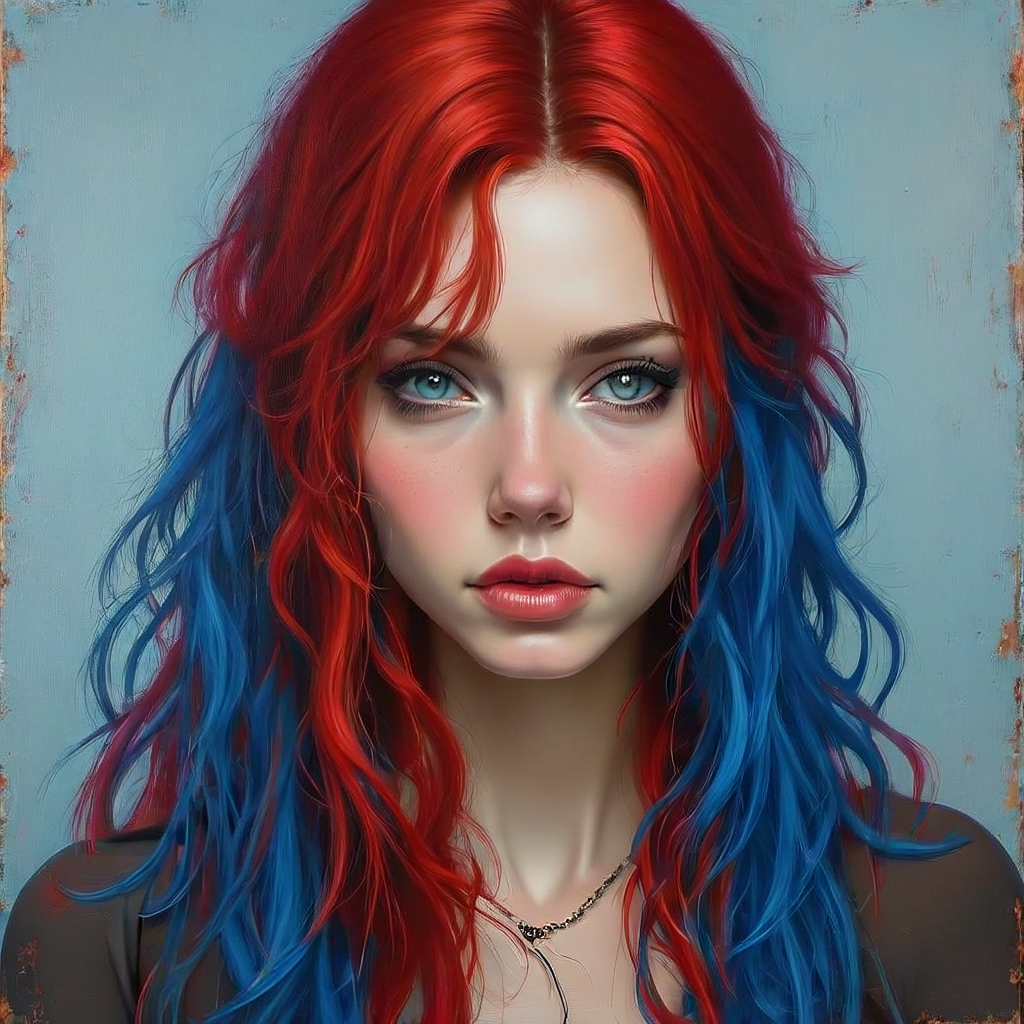}
    \end{tabular}
    \caption{Infinity-2B~\cite{han2025infinity} with \ourmethod at 10\% budget}
    \label{fig:image_collage_2b}
\end{figure}

\begin{figure}[!ht]
    \centering
    \setlength{\tabcolsep}{1pt}
    \renewcommand{\arraystretch}{0}
    \begin{tabular}{ccc}
        \includegraphics[width=0.328\linewidth]{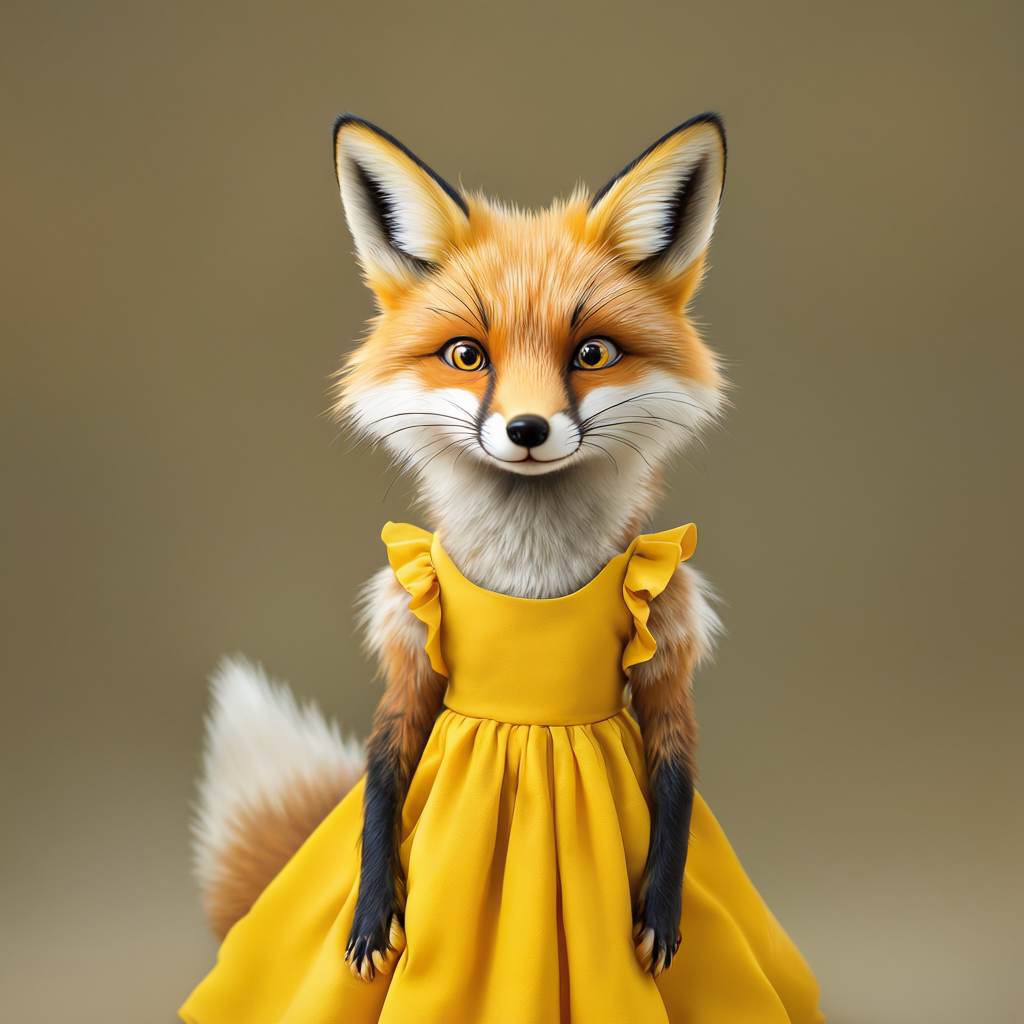} &
        \includegraphics[width=0.328\linewidth]{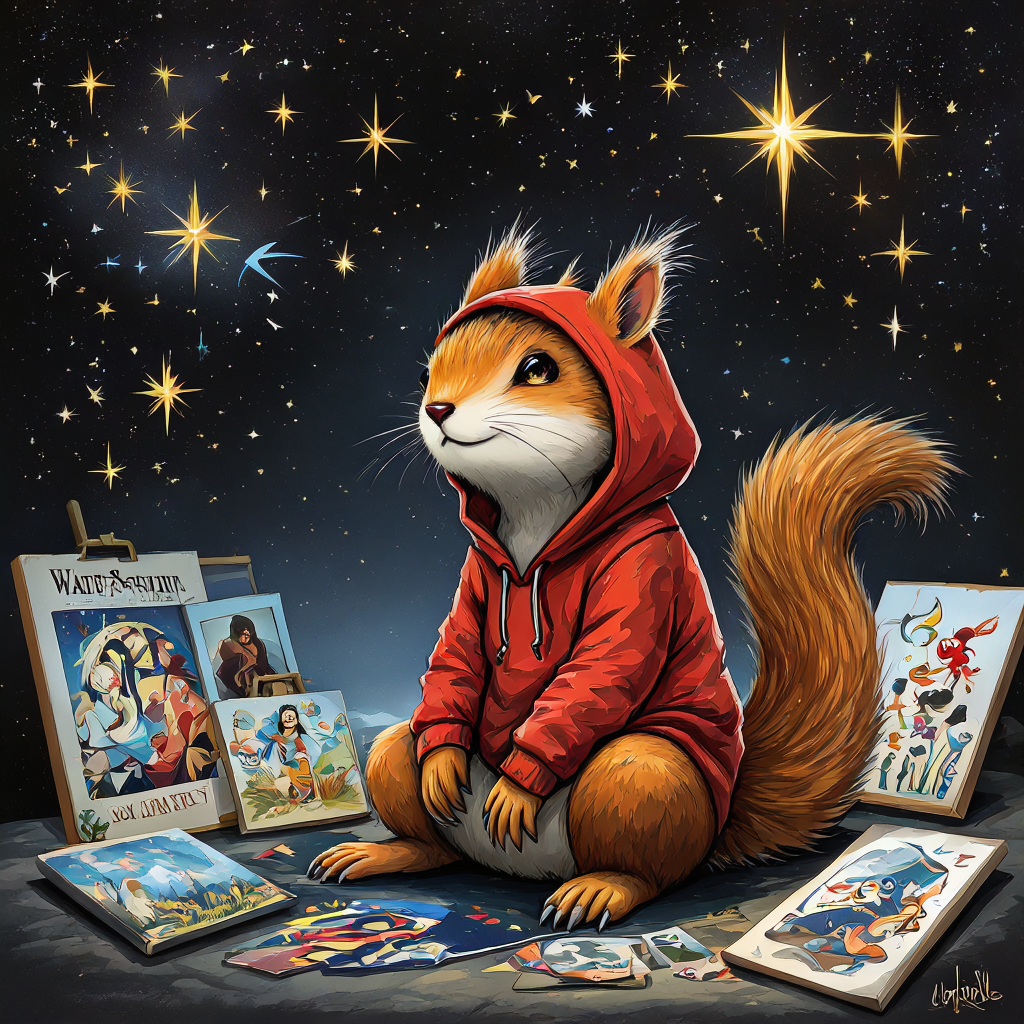} &
        \includegraphics[width=0.328\linewidth]{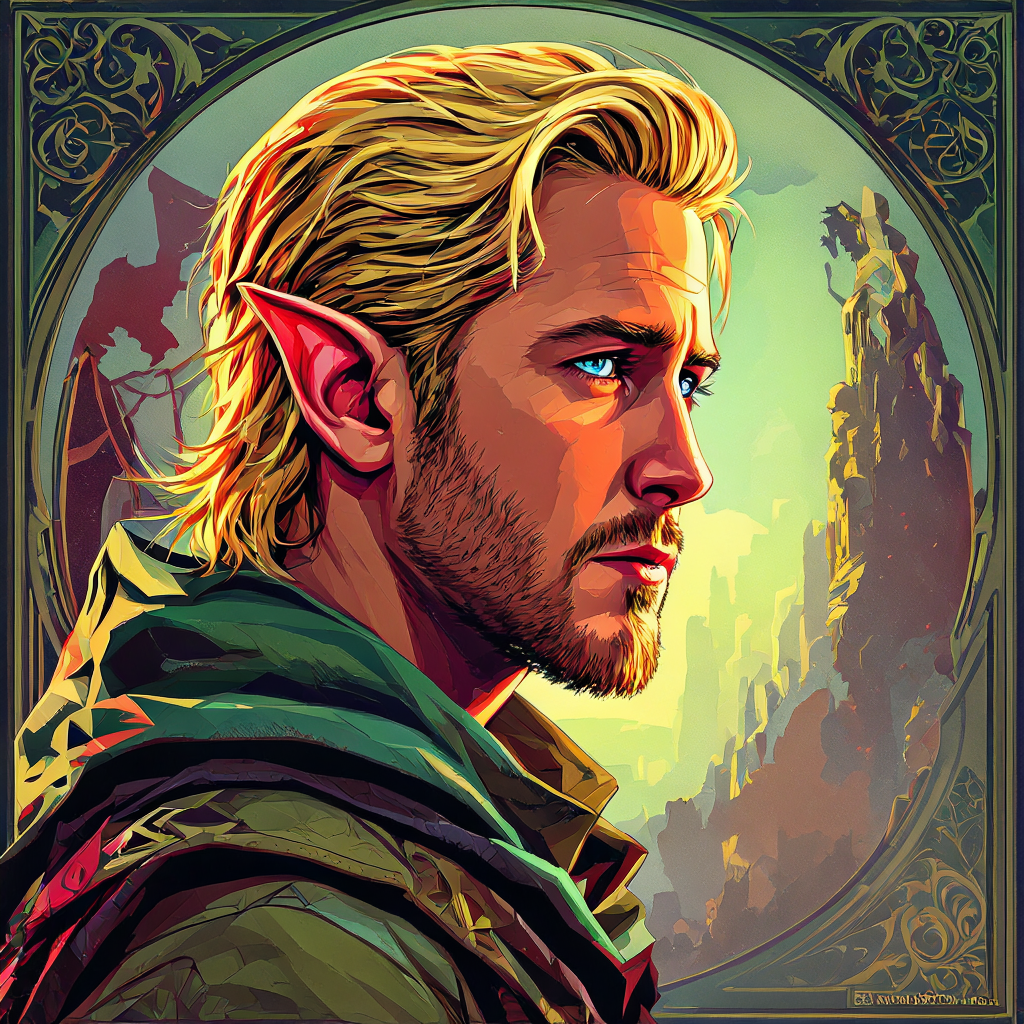} \\
        \includegraphics[width=0.328\linewidth]{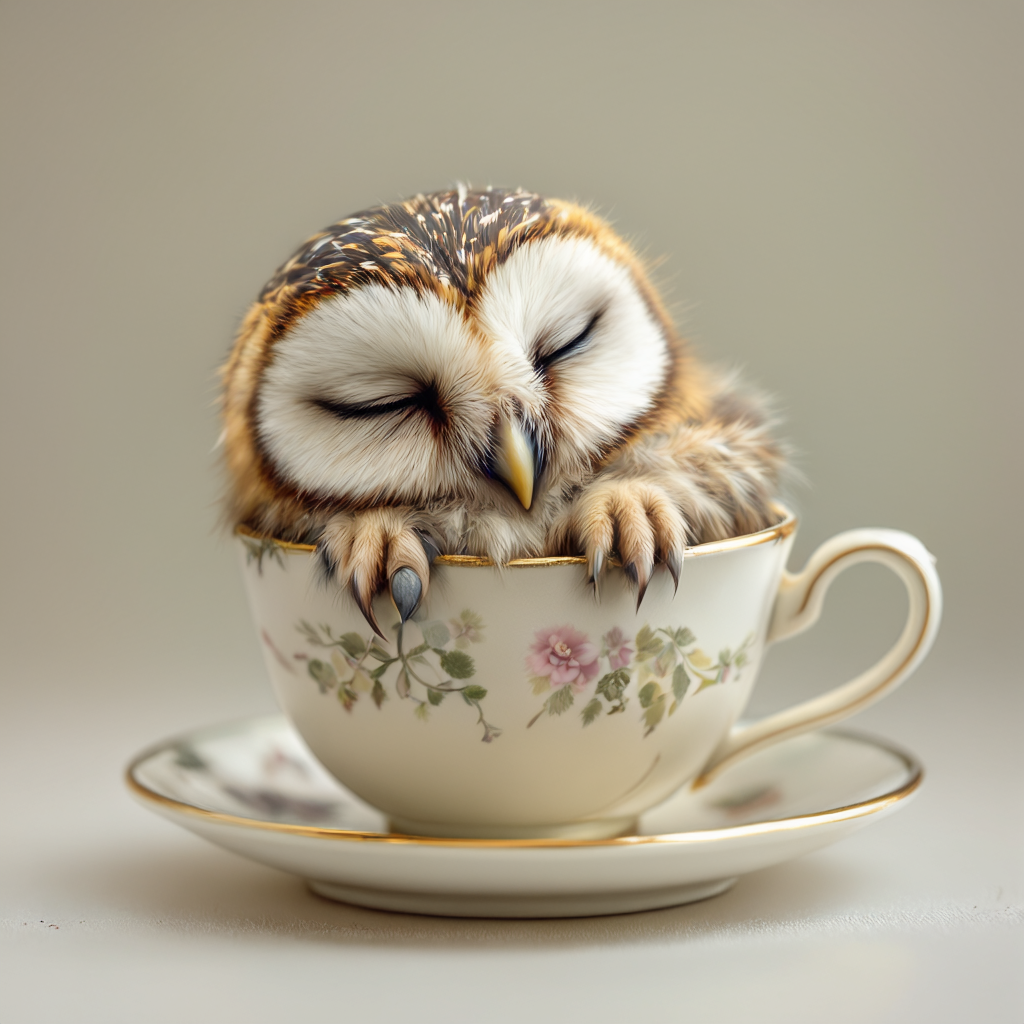} &
        \includegraphics[width=0.328\linewidth]{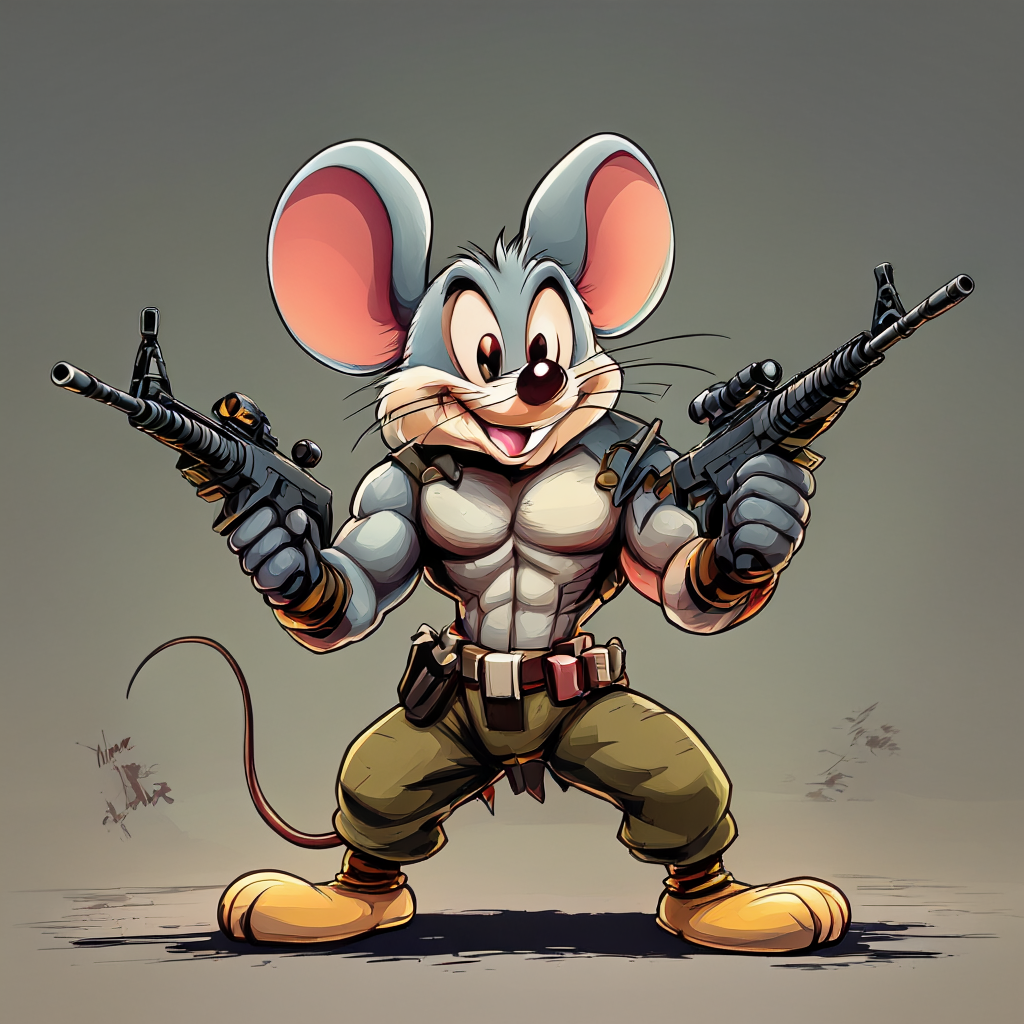} &
        \includegraphics[width=0.328\linewidth]{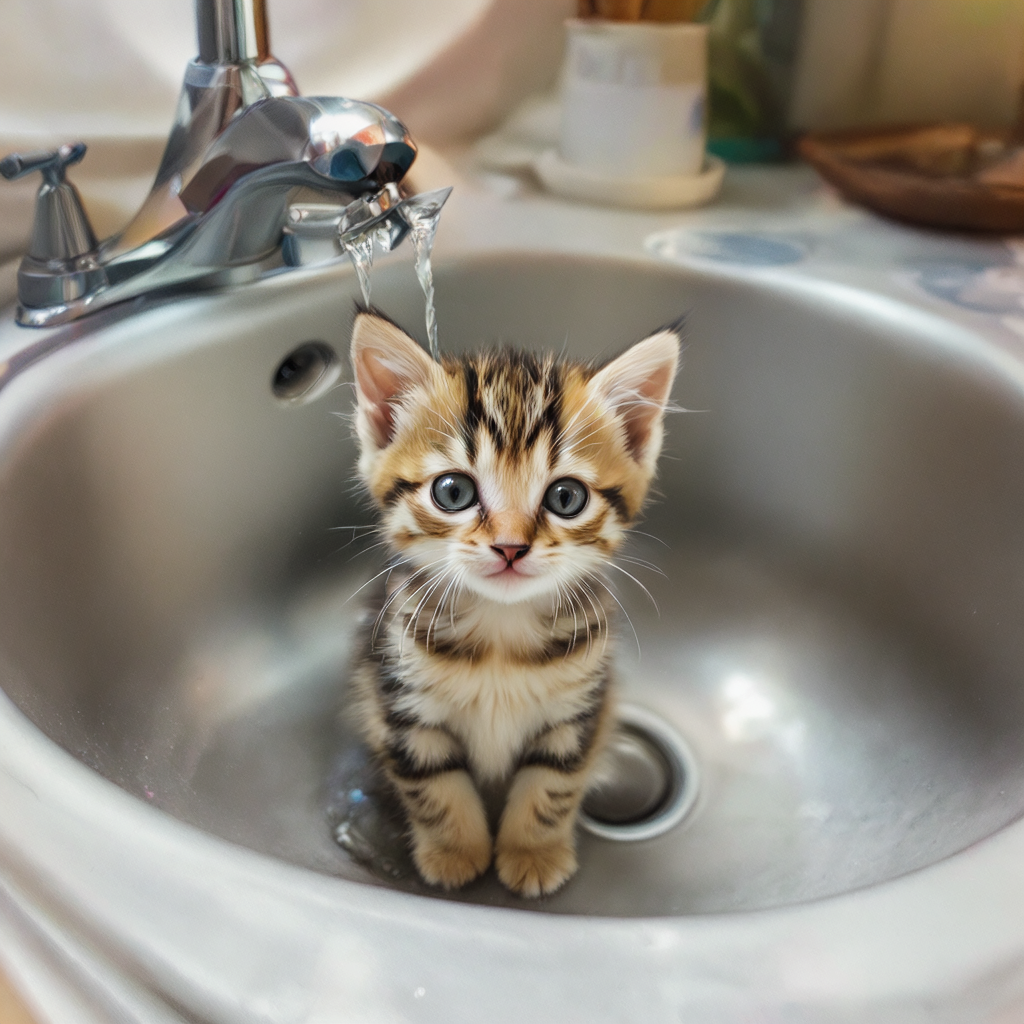} \\
        \includegraphics[width=0.328\linewidth]{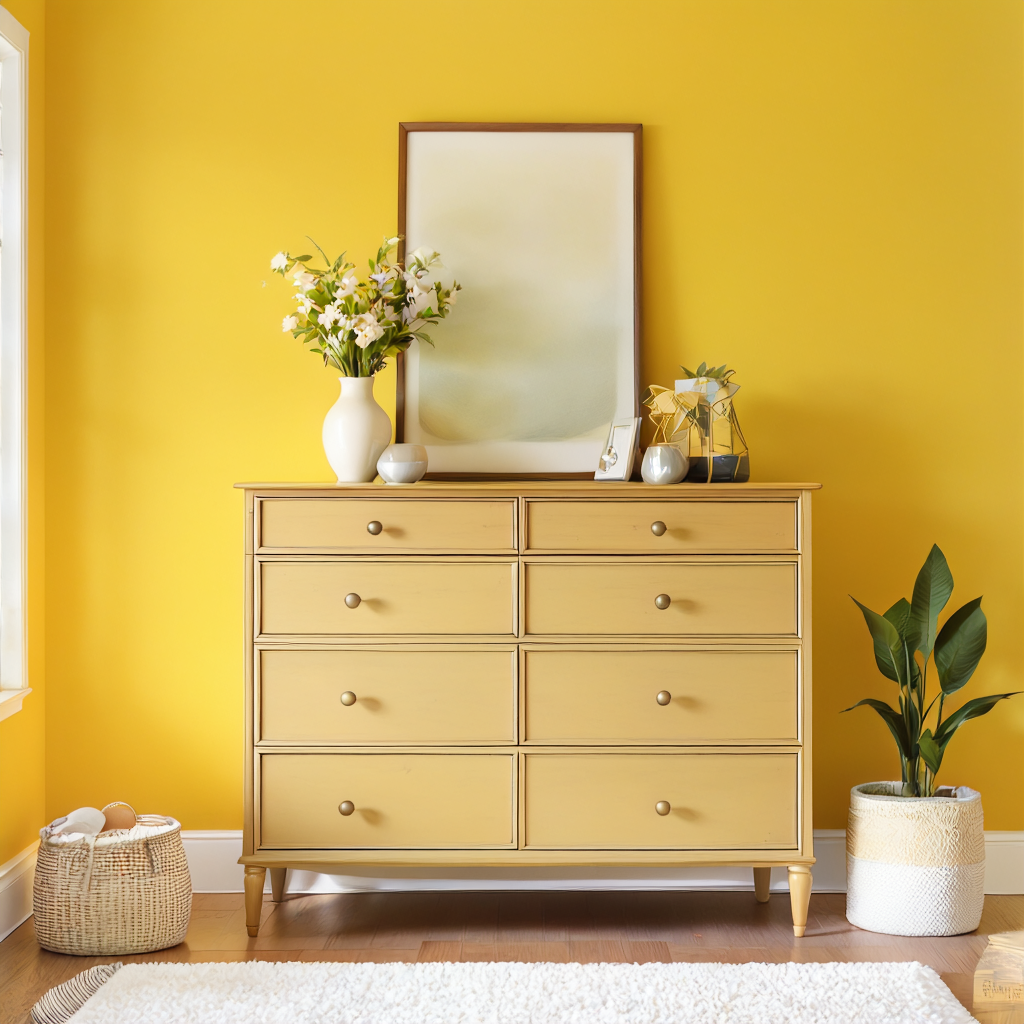} &
        \includegraphics[width=0.328\linewidth]{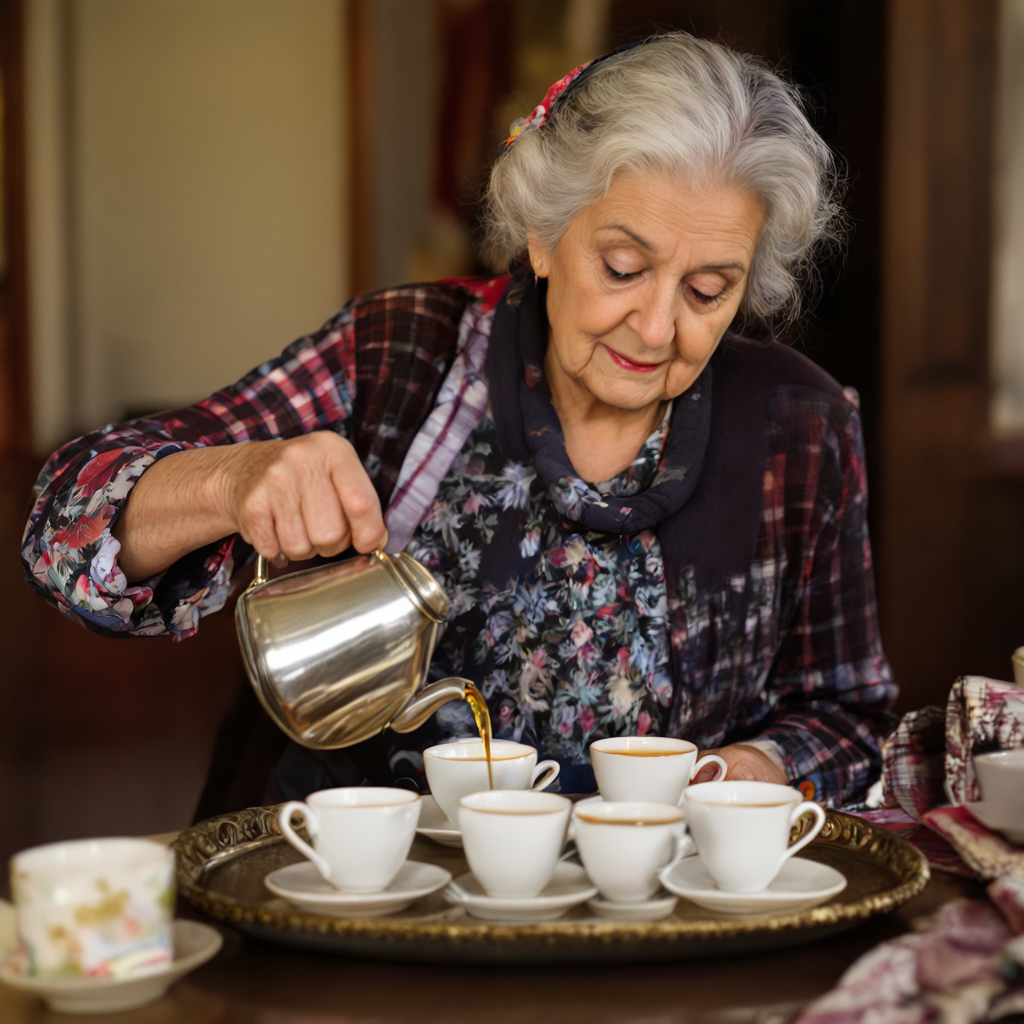} &
        \includegraphics[width=0.328\linewidth]{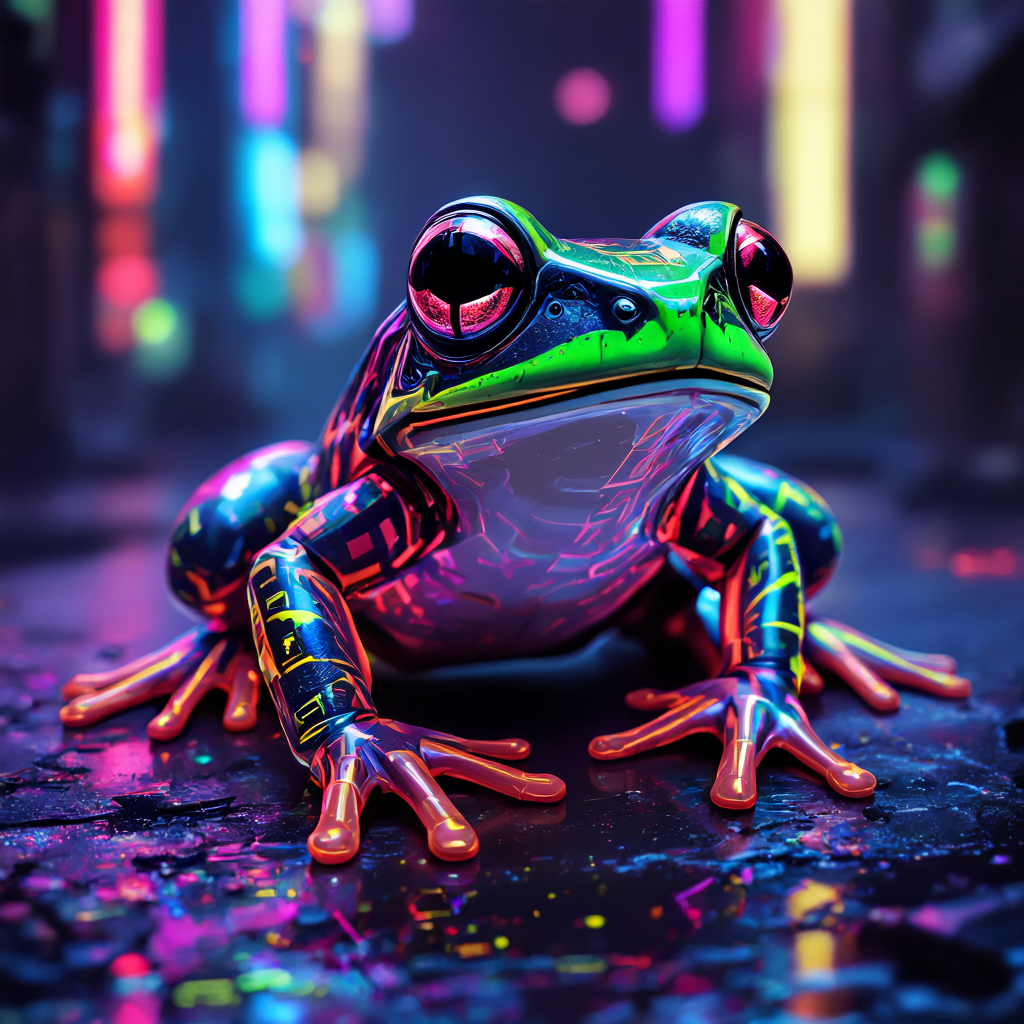}
    \end{tabular}
    \caption{Infinity-8B~\cite{han2025infinity} with \ourmethod at 10\% budget}
    \label{fig:image_collage_8b}
\end{figure}

\end{document}